\documentclass[10pt,twocolumn,letterpaper]{article}

\usepackage[pagenumbers]{cvpr} 
\usepackage{booktabs}
\usepackage{multirow}
\usepackage[table]{xcolor}
\usepackage{tabularx}
\usepackage{array}

\usepackage{algorithm}
\usepackage{algpseudocode}

\definecolor{cvprblue}{rgb}{0.21,0.49,0.74}
\usepackage[pagebackref,breaklinks,colorlinks,allcolors=cvprblue]{hyperref}

\title{Decision Boundaries Matter: Graph-Regularized Residual Adaptation for Cross-Domain Fake Image Detection}

\author{
Shichuang Xie$^{1}$ \quad Qishuai Ding$^{1}$ \quad Tong Qiao$^{2}$ \quad Liquan Shen$^{1}$ \quad Guorui Feng$^{1,*}$ \\[0.5em]
$^{1}$School of Communication Information Engineering, Shanghai University, Shanghai, China \\
$^{2}$School of Cyberspace, Hangzhou Dianzi University, Hangzhou, China \\[0.5em]
{\tt\small *Corresponding author: grfeng@shu.edu.cn}
}

\begin{document}
\maketitle
\begin{abstract}

Despite learning transferable features, fake image detectors may struggle on unseen domains due to mismatched decision boundaries. Our analysis shows that features extracted by the frozen forensic encoder remain discriminative, but the source-trained decision boundary does not fully exploit this information. However, correcting this boundary without source data or target labels remains challenging, as unreliable source predictions may lead to incorrect adaptation. To address this issue, we propose Graph-Regularized Residual Adaptation (GraRA), which refines the source decision boundary using only unlabeled target data. To determine which classifier directions to update, GraRA constructs a compact residual subspace from uncertainty-weighted target features and keeps it orthogonal to the source classification direction. To guide these updates, it uses source-score anchors to generate soft labels through target-graph diffusion, while neighborhood consistency regularizes residual learning within the constrained subspace. The learned correction is merged into the source classifier, requiring only 32 optimized coefficients and no additional inference branches. On diverse unseen OpenMMSec domains, GraRA achieves 84.36\% accuracy and 91.34\% AUC, improving over the source detector by 4.33 and 3.50 percentage points, respectively. Further analysis shows that reliable source predictions and locally consistent neighborhoods help guide residual learning, allowing GraRA to correct ranking errors that a simple threshold shift cannot address. \textcolor{cvprblue}{Code will be made publicly available.}

\end{abstract}

\section{Introduction}
\label{sec:intro}

\begin{figure}[!htp]
    \centering
    \includegraphics[width=\linewidth]{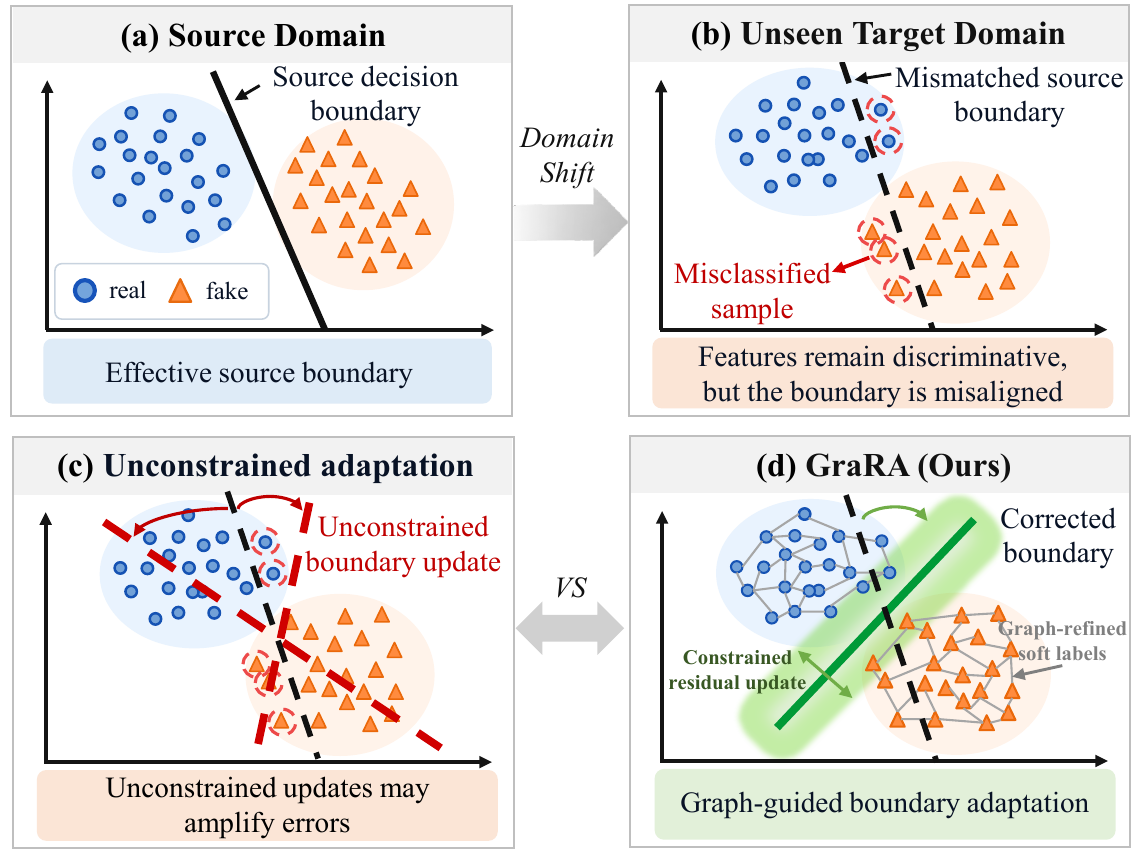}
    \caption{\textbf{Motivation for GraRA.}
    (a) The source classifier separates real and fake samples in the source domain.
    (b) Domain shifts can cause decision boundary mismatch even when frozen features remain discriminative.
    (c) Unconstrained adaptation may amplify errors from unreliable source predictions.
    (d) GraRA combines graph-refined supervision with constrained residual learning to correct the source decision boundary.}
    \label{fig:teaser}
    \vspace{-0.3cm}
\end{figure}

\begin{figure*}[!htp]
    \centering
    \includegraphics[width=\linewidth]{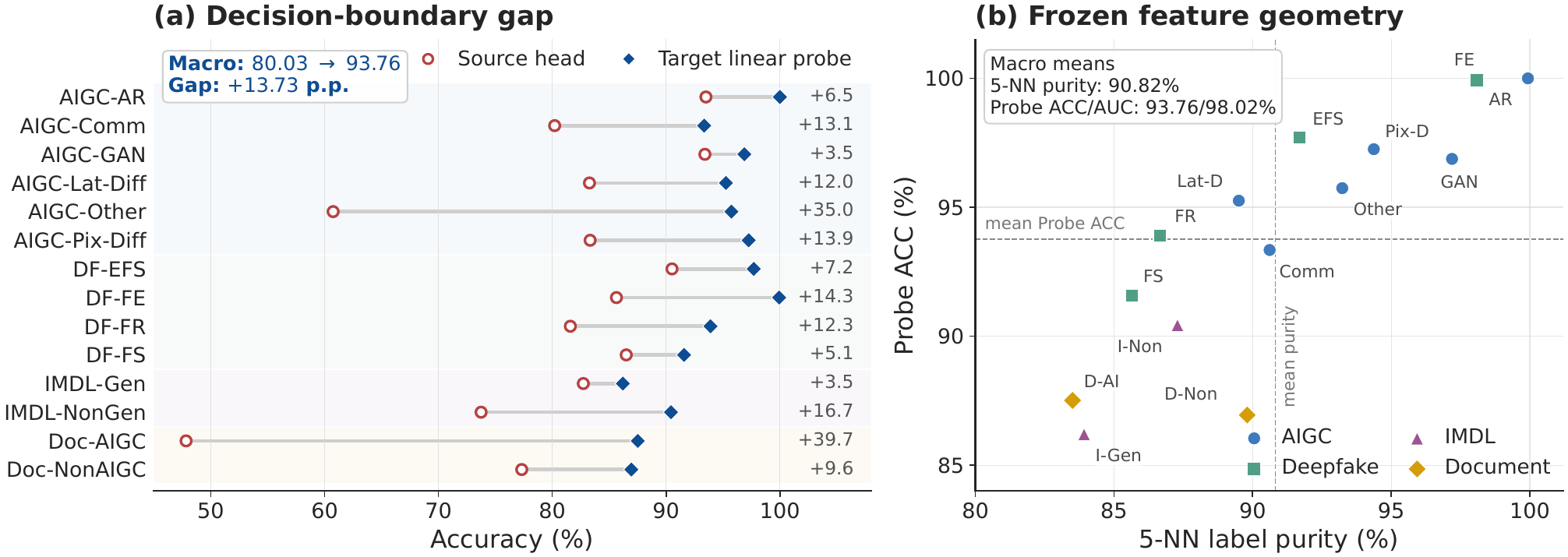}
    \caption{\textbf{Evidence of decision boundary mismatch across diverse target domains.}
    (a) A target-supervised linear probe improves accuracy over the frozen source classifier, showing how much target discrimination can be recovered by changing only the classifier.
    (b) 5-NN label purity and linear-probe performance measure the local class structure and linear separability preserved in frozen target features.}
    \label{fig:motivation}
    \vspace{-0.4cm}
\end{figure*}

In real-world forensic scenarios, fake images can be produced by diverse and evolving forgery techniques, including new face-swapping methods~\cite{blendface,e4s,li2024learning}, GAN-based generators~\cite{stylegan-xl,NEURIPS2021_076ccd93}, and diffusion models~\cite{Rombach2021HighResolutionIS,dit}. Image editing, compression, and post-processing can further change or obscure traces of manipulation. As a result, detectors trained on known forgeries may struggle to generalize to unseen manipulation methods and processing conditions. Maintaining reliable detection under such distribution shifts remains a key challenge in image forensics.

Existing studies~\cite{Yan2023UCFUC,yan2024transcending,wang2023dire,chen2024enhancing,Jing2026ResProtoFDVR} improve cross-domain generalization by learning transferable forensic features. Recent methods build on vision foundation models~\cite{awais2025foundation}, including ForAda~\cite{cui2025forensics}, EFFORT~\cite{yan2025orthogonal}, and SICA~\cite{du2026sica}, which use forensic adapters, orthogonal subspaces, and semantic-guided adaptation, respectively. These methods aim to encode useful pretrained knowledge while learning discriminative forensic features. However, detection performance also depends on the source decision boundary. Even when frozen features preserve useful information for distinguishing real and fake target images, the source classifier may fail to use this information effectively. This raises a question: \emph{How can we improve cross-domain fake image detection by adapting only the decision boundary?}

To understand this issue, consider how a binary fake image detector makes predictions. Its forensic encoder extracts cues such as texture patterns, synthesis artifacts, and local inconsistencies. A source-trained classifier then maps these features to real and fake scores. However, unseen generators may change the distribution and relative strength of these cues, making the source decision boundary unsuitable for target images. As shown in Figure~\ref{fig:teaser}(a) and (b), the source classifier can separate real and fake images in the source domain but misclassify target samples when the decision boundary no longer matches the target distribution. This mismatch suggests that the source decision boundary may lead to poor generalization on target domains, even when the encoder preserves useful forensic cues.

To test this hypothesis, we consider two questions. \textbf{Q1:} Can a target-supervised linear classifier recover discriminative performance with the encoder frozen? \textbf{Q2:} Do frozen target features preserve useful local forensic structure? To answer them, we conduct two diagnostic experiments on diverse target domains from the OpenMMSec benchmark~\cite{du2026sica}. Details appear in the supplementary material (Sec.~B.1-B.2). We train a target-supervised linear classifier on frozen features and examine their local label consistency. As shown in Figure~\ref{fig:motivation}, the linear classifier achieves 93.76\% accuracy and 98.02\% AUC, improving accuracy over the source classifier by 13.73 percentage points. The frozen features also achieve an average 5-NN label purity of 90.82\%. These results show that frozen target features encode discriminative information and local structure that the source classifier does not fully exploit.

However, the supervised linear probe in Figure~\ref{fig:motivation}(a) requires target labels, which are unavailable during adaptation. We consider a source-free setting with only a pretrained detector and unlabeled target data. This setting applies to real-world forensic systems, where source training data may be unavailable due to privacy concerns or limited data access. Adaptation methods such as IST~\cite{ma2024improved} and GOLD~\cite{lai2026golden} refine pseudo-labels or perform low-rank feature adaptation. FPS~\cite{cheng2026feature} adapts decision boundaries using frozen features but requires labeled source features. Yet refining pseudo-labels alone does not constrain boundary updates, while restricting update directions does not ensure reliable target supervision. This challenge is further complicated by unreliable source predictions and neighborhoods containing samples from different classes on unseen target domains. Direct adaptation with such supervision may amplify errors, as shown in Figure~\ref{fig:teaser}(c). These challenges raise three design questions: (1) which classifier directions to update, (2) how to refine unreliable source predictions using target features, and (3) how to keep boundary updates consistent with local feature structure.

To address these questions, we propose Graph-Regularized Residual Adaptation (GraRA), a source-free method that adapts a pretrained detector's decision boundary. As illustrated in Figure~\ref{fig:teaser}(d), GraRA constructs a compact residual subspace from uncertainty-weighted target features and keeps it orthogonal to the source classification direction. It then uses source-score anchors to guide semantic diffusion on a target $k$-NN graph, generating soft labels for adaptation. These soft labels and graph consistency guide residual optimization within the constrained subspace. Thus, the subspace restricts update directions, while the graph provides both semantic supervision and structural regularization. The learned correction is merged into the source classifier without additional inference branches. Our contributions are as follows:

\begin{itemize}
    \item We revisit cross-domain fake image detection from the perspective of decision-boundary mismatch. Our analysis shows that the source-trained boundary does not fully exploit the discriminative information captured by the frozen forensic encoder.
    
    \item We propose GraRA to adapt the source decision boundary using only unlabeled target images. It refines supervision from source predictions using local feature structure and constrains classifier updates through a compact source-orthogonal residual and graph consistency.
    
    \item Extensive comparisons on diverse domains demonstrate GraRA's effectiveness. Further analyses examine the contributions of its residual learning components.
\end{itemize}

\section{Related Work}
\label{sec:related}

\subsection{Generalizable Fake Image Detection}
Fake image detection~\cite{du2026forensichub} covers four common settings: Deepfake detection for facial forgeries \cite{xie2025deepfake}, AI-generated content (AIGC) detection for fully synthesized images, image manipulation detection and localization (IMDL) for partially edited natural images, and document forgery detection (Doc) for manipulated documents. Existing methods rely on domain-specific forensic cues. Deepfake detectors exploit facial artifacts, frequency cues, or pretrained features~\cite{NEURIPS2023_0e735e4b, shiohara2022detecting,hanzhefreqblender,qiao2024fully, cui2025forensics}. 
AIGC detectors exploit generation artifacts and pretrained representations~\cite{wang2023dire,ojha2023towards,yan2025orthogonal,dhakal2026simlbr, Yang_2026_CVPR}. IMDL methods target local inconsistencies~\cite{guillaro2023trufor,ma2024imdl,zhu2025mesoscopic}, while document forgery detectors analyze text morphology and spatial--frequency cues~\cite{qu2023towards,chen2024enhancing}. However, methods designed for specific forgery types may generalize poorly across forensic domains.

Recent studies explore unified detection frameworks. ForensicHub~\cite{du2026forensichub} provides a benchmark covering diverse forensic domains, while UniShield~\cite{huang2026unishield} combines specialized experts for unified detection. UnivFD~\cite{ojha2023towards} demonstrates the transferability of pretrained representations across generators. Among recent state-of-the-art (SOTA) approaches, EFFORT~\cite{yan2025orthogonal} decomposes pretrained features into orthogonal subspaces to preserve general knowledge while learning forgery-specific cues. SICA~\cite{du2026sica} uses semantic-guided constrained adaptation to construct a unified yet discriminative forensic feature space across forgery types. These methods focus on learning transferable representations during training. In contrast, we revisit cross-domain fake image detection from a complementary perspective. We adapt the decision boundary of a trained detector to unseen domains using only unlabeled target images, while keeping the feature encoder frozen.

\subsection{Source-Free and Frozen-Feature Adaptation}

Source-free domain adaptation and test-time adaptation aim to adapt pretrained models using unlabeled target data without accessing source training samples. SHOT~\cite{liang2020we} freezes the source classifier and adapts the feature extractor through self-training, while TENT~\cite{wang2021tent}, EATA~\cite{niu2022efficient}, and CoTTA~\cite{wang2022continual} update model parameters to address distribution shifts. Recent methods further improve the reliability of target supervision. DIFO~\cite{tang2024source} uses a vision--language model to guide source-free adaptation through knowledge distillation, while IST~\cite{ma2024improved} exploits feature similarity graphs to refine pseudo-labels. Although these methods avoid access to source data, they typically update model parameters to adapt to the target distribution. However, such updates may be unnecessary when frozen features already contain useful discriminative information for the target domain.


Other studies explore adaptation with frozen features. FreeTTA~\cite{dai2025free} models target feature distributions through online expectation-maximization without gradient-based updates. GOLD~\cite{lai2026golden} projects frozen features onto a low-rank subspace and learns lightweight scaling parameters for continual test-time adaptation. FPS~\cite{cheng2026feature} searches for decision planes in a frozen feature space, but relies on labeled source data and multiple strategies to stabilize boundary adaptation. In contrast, GraRA uses local structure in unlabeled target features to refine source-guided supervision, while combining a compact residual with graph consistency to constrain boundary adaptation and improve sample ranking.

\begin{figure*}[!thp]
    \centering
    \includegraphics[width=\linewidth]{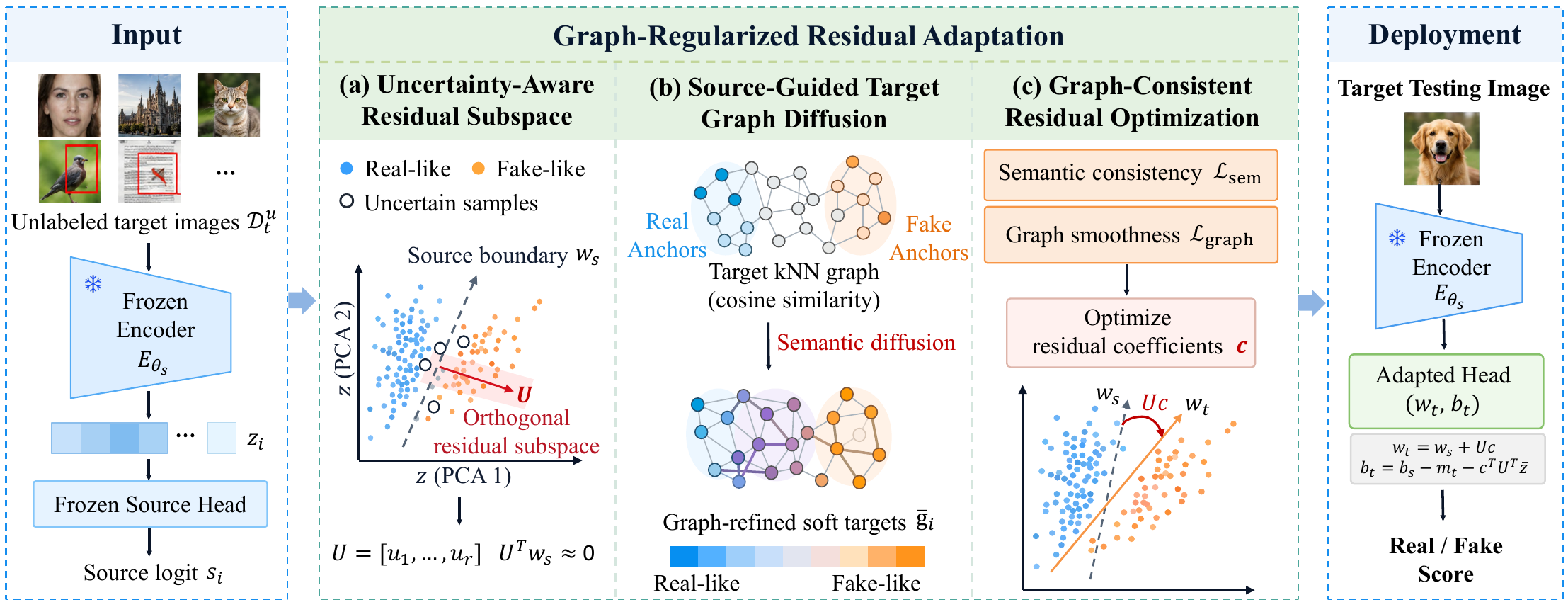}
    \caption{\textbf{Overview of GraRA.} Given unlabeled target images, a frozen source detector extracts target features and source logits. GraRA then adapts the source classifier in three stages. (a) An \textbf{Uncertainty-Aware Residual Subspace} is constructed from uncertainty-weighted target features and kept orthogonal to the source classification direction to constrain classifier updates. (b) \textbf{Source-Guided Target Graph Diffusion} uses real and fake anchors to guide semantic diffusion on a target $k$-NN graph, generating graph-refined soft targets. (c) \textbf{Graph-Consistent Residual Optimization} combines semantic consistency and graph smoothness losses to optimize residual coefficients within the constrained subspace. The learned residual is merged into the source head for target inference.}
    \label{fig:overview}
\end{figure*}

\section{Graph-Regularized Residual Adaptation}
\label{sec:method}

\subsection{Problem Formulation and Overview}
\label{sec:setting}

Consider a binary fake image detector trained on a labeled source domain. It consists of a feature encoder $E_{\boldsymbol{\theta}_s}$ and a linear classifier $(\boldsymbol{w}_s,b_s)$. At deployment, we have access to an unlabeled target adaptation batch $D_t^u=\{\boldsymbol{x}_i\}_{i=1}^{N_t}$. The detector extracts a feature vector and computes a classification logit for each image:
\begin{equation}
    \boldsymbol{z}_i=E_{\boldsymbol{\theta}_s}(\boldsymbol{x}_i)\in\mathbb{R}^d,
    \qquad
    s_i=\boldsymbol{w}_s^{\top}\boldsymbol{z}_i+b_s,
    \label{eq:source_logit}
\end{equation}
where higher logits indicate stronger evidence of forgery. The source detector, including its encoder and classifier, remains frozen throughout adaptation. Our goal is to adapt the linear classifier to the target domain using only unlabeled target images, without access to source samples.

Although frozen target features may preserve useful discriminative information, the source decision boundary may not fit unseen target domains. GraRA addresses this mismatch through constrained residual optimization guided by target feature relationships. It first constructs an uncertainty-aware residual subspace orthogonal to the source classification direction, restricting update directions. Source-score anchors then guide graph diffusion to generate soft labels from target neighborhoods. These labels and graph consistency jointly guide residual optimization within the subspace. Figure~\ref{fig:overview} illustrates the overall framework.

Before residual learning, we select the samples with the lowest and highest source logits as anchors, using an anchor selection ratio $\zeta$ for each group. The two anchor sets are denoted by $\mathcal{A}_{\mathrm r}$ and $\mathcal{A}_{\mathrm f}$, respectively. Their median scores are averaged to obtain a reference score $m_t$. Since the source decision threshold may not suit the target domain, we use $m_t$ to center the logits between the two anchor groups. We then compute the shifted logit $\widetilde{s}_i=s_i-m_t$ and the corresponding predicted probability $\widetilde{p}_i=\sigma(\widetilde{s}_i)$, where $\sigma(\cdot)$ is the sigmoid function. These shifted probabilities are used for uncertainty weighting and graph diffusion.

\subsection{Uncertainty-Aware Residual Subspace}
\label{sec:residual_geometry}

Directly optimizing all classifier weights without target labels may lead to unreliable updates~\cite{niu2022efficient}. To reduce this risk, we construct a compact residual subspace from uncertainty-weighted target features. The subspace captures target feature variations near the shifted source boundary and restricts classifier updates to directions orthogonal to the source classification direction. We first assign higher weights to target samples with uncertain predictions:
\begin{equation}
    q_i=\max\!\left\{\epsilon,\,
        4\widetilde{p}_i(1-\widetilde{p}_i)\right\},
    \label{eq:calibrated_uncertainty}
\end{equation}
where $\epsilon=10^{-3}$ prevents zero weights. The weight reaches its maximum when $\widetilde{p}_i=0.5$ and decreases as the prediction becomes more confident. This allows us to focus on samples near the decision boundary, where classification errors are more sensitive to boundary changes.

We use these weights to compute the weighted mean and covariance of the target features:
\begin{equation}
\begin{aligned}
    \boldsymbol{\mu}_q
    &=\frac{\sum_{i=1}^{N_t}q_i\boldsymbol{z}_i}
            {\sum_{i=1}^{N_t}q_i},\\
    \boldsymbol{C}_q
    &=\frac{\sum_{i=1}^{N_t}q_i
      (\boldsymbol{z}_i-\boldsymbol{\mu}_q)(\boldsymbol{z}_i-\boldsymbol{\mu}_q)^{\top}}
      {\sum_{i=1}^{N_t}q_i}.
\end{aligned}
\label{eq:weighted_covariance}
\end{equation}
We then apply principal component analysis (PCA) to the weighted covariance matrix $\boldsymbol{C}_q$ to identify the main directions of target feature variation, giving greater weight to uncertain samples. We select the top $r+\delta$ eigenvectors as candidate directions for residual adaptation:
\begin{equation}
\begin{aligned}
    \boldsymbol{C}_q\mathbf{a}_j &=\gamma_j\mathbf{a}_j,
       \quad \gamma_1\geq\cdots\geq\gamma_{r+\delta},\\
    \boldsymbol{A}_{r+\delta} &= [\mathbf{a}_1,\ldots,\mathbf{a}_{r+\delta}].
\end{aligned}
    \label{eq:pca_candidates}
\end{equation}
Here, $r$ is the desired residual dimension, and $\delta$ provides extra candidate directions in case of rank loss during the subsequent projection.

We construct the residual basis by removing the component of each PCA direction that is parallel to the source classifier weight:
\begin{equation}
    \bar{\boldsymbol{w}}_s=\frac{\boldsymbol{w}_s}{\lVert\boldsymbol{w}_s\rVert_2},
    \qquad
    \boldsymbol{U}=\operatorname{orth}_r
    \left((\boldsymbol{I}-\bar{\boldsymbol{w}}_s\bar{\boldsymbol{w}}_s^{\top})\boldsymbol{A}_{r+\delta}\right).
    \label{eq:residual_basis}
\end{equation}
Here, $\operatorname{orth}_r(\cdot)$ selects and orthonormalizes the first $r$ independent directions. If at least $r$ independent directions remain after projection, the basis satisfies $\boldsymbol{U}^{\top}\boldsymbol{U}=\boldsymbol{I}_r$ and $\boldsymbol{U}^{\top}\bar{\boldsymbol{w}}_s=\boldsymbol{0}$. During adaptation, we keep $\boldsymbol{U}$ fixed and optimize only $r$ residual coefficients, restricting weight updates to directions orthogonal to the source classifier.

\subsection{Source-Guided Target Graph Diffusion}
\label{sec:graph}

The residual basis $\boldsymbol{U}$ defines the directions available for classifier adaptation, but it does not determine how these directions should be combined. Since target labels are unavailable, we use the source classifier's predictions to initialize diffusion on a graph constructed from frozen target features. The resulting soft labels guide residual learning.

\noindent\textbf{Target Similarity Graph.}
We first construct a directed $k$-NN graph from the frozen target features. Each feature is normalized as $\widehat{\boldsymbol{z}}_i=\boldsymbol{z}_i/\lVert\boldsymbol{z}_i\rVert_2$, and the $k$ most similar samples are selected using cosine similarity. Let $\mathcal{N}_k(i)$ denote the neighbors of sample $i$. We compute the edge weights as
\begin{equation}
    a_{ij}
    =\frac{\exp(\widehat{\boldsymbol{z}}_i^{\top}\widehat{\boldsymbol{z}}_j/\tau)}
    {\sum_{l\in\mathcal{N}_k(i)}
     \exp(\widehat{\boldsymbol{z}}_i^{\top}\widehat{\boldsymbol{z}}_l/\tau)},
    \quad j\in\mathcal{N}_k(i),
    \label{eq:graph_affinity}
\end{equation}
where $\tau>0$ controls how strongly the weights favor more similar neighbors. These weights control each neighbor's influence in semantic diffusion and residual optimization.

\noindent\textbf{Anchor Initialization.}
We initialize the graph probabilities $g_i^{(0)}$ using the anchor sets $\mathcal{A}_{\mathrm r}$ and $\mathcal{A}_{\mathrm f}$ defined in Sec.~\ref{sec:setting}. Specifically, $g_i^{(0)}$ is set to 0 for real anchors and 1 for fake anchors, while the remaining samples use their predicted probabilities $\widetilde{p}_i$. The anchors are selected without target labels and may contain prediction errors.

\noindent\textbf{Confidence-Guided Semantic Diffusion.}
We update each sample's probability by combining its initial prediction with the weighted predictions of its neighbors:
\begin{equation}
    g_i^{(t+1)}
    =\lambda_i g_i^{(0)}
    +(1-\lambda_i)
      \sum_{j\in\mathcal{N}_k(i)}a_{ij}g_j^{(t)}.
    \label{eq:diffusion}
\end{equation}
Here, $\lambda_i$ controls how much the updated probability relies on the initial source prediction. For non-anchor samples, we define
\begin{equation}
    \lambda_i
    =\lambda_{\mathrm{weak}}
    +(\lambda_{\mathrm{anc}}-\lambda_{\mathrm{weak}})
      \lvert2\widetilde p_i-1\rvert,
    \label{eq:confidence_tether}
\end{equation}
where $0<\lambda_{\mathrm{weak}}<\lambda_{\mathrm{anc}}<1$. For anchor samples, we set $\lambda_i=\lambda_{\mathrm{anc}}$. Thus, uncertain samples rely more on their neighbors, while confident samples retain more information from their initial predictions.

After $T$ diffusion steps, we use the resulting probabilities $\bar g_i=g_i^{(T)}$ as soft labels for residual learning. The same graph is also used to encourage consistent predictions between neighboring target samples.

\subsection{Graph-Consistent Residual Learning}
\label{sec:residual_learning}

Given the constrained residual basis and graph-diffused soft labels, we learn a compact correction to the source decision boundary. Soft-label supervision and neighborhood consistency jointly guide residual optimization.

We first compute the mean target feature
$\bar{\boldsymbol{z}}$ and project each centered
feature onto the fixed residual basis
$\boldsymbol{U}$ to obtain its residual
coordinates $\boldsymbol{h}_i$:
\begin{equation}
    \bar{\boldsymbol{z}}=\frac{1}{N_t}\sum_{i=1}^{N_t}\boldsymbol{z}_i,
    \qquad
    \boldsymbol{h}_i=\boldsymbol{U}^{\top}(\boldsymbol{z}_i-\bar{\boldsymbol{z}}).
    \label{eq:centered_coordinate}
\end{equation}
We then introduce a learnable residual coefficient
vector $\boldsymbol{c}\in\mathbb{R}^r$ to adjust
the classification direction.
The adapted logit $\ell_i$ is computed as
\begin{equation}
    \ell_i=\widetilde{s}_i+\boldsymbol{c}^{\top}\boldsymbol{h}_i,
    \label{eq:adapted_logit}
\end{equation}
where $\widetilde{s}_i$ is the shifted source
logit and $\boldsymbol{c}^{\top}\boldsymbol{h}_i$
is the residual correction.
Since $\sum_i\boldsymbol{h}_i=\boldsymbol{0}$,
the residual correction has zero mean over
the target samples. Thus, the reference shift
$m_t$ remains fixed, while the learned residual
adjusts the classification direction.

The coefficient vector $\boldsymbol{c}$ is optimized using two losses. The first is a soft-label cross-entropy loss $\mathcal{L}_{\mathrm{sem}}$ that aligns adapted predictions with graph-diffused probabilities:
\begin{equation}
    \mathcal{L}_{\mathrm{sem}}
    =-\frac{1}{N_t}\sum_{i=1}^{N_t}
      \left[
        \bar g_i\log\sigma(\ell_i)
        +(1-\bar g_i)\log\bigl(1-\sigma(\ell_i)\bigr)
      \right].
    \label{eq:semantic_loss}
\end{equation}
Here, $\bar g_i$ is the soft label obtained in Sec.~\ref{sec:graph}. These soft labels remain fixed during residual optimization.

The second loss $\mathcal{L}_{\mathrm{graph}}$ encourages similar target samples to receive similar predictions. Using the target graph constructed in Sec.~\ref{sec:graph}, we define
\begin{equation}
    \mathcal{L}_{\mathrm{graph}}
    =\frac{1}{N_t}\sum_{i=1}^{N_t}
      \sum_{j\in\mathcal{N}_k(i)}
      a_{ij}\bigl[\sigma(\ell_i)-\sigma(\ell_j)\bigr]^2.
    \label{eq:graph_loss}
\end{equation}
This loss penalizes large prediction differences between neighboring samples, with higher weights assigned to more similar neighbors. The final objective combines both losses:
\begin{equation}
    \mathcal{L}_{\mathrm{GraRA}}
    =(1-\alpha)\mathcal{L}_{\mathrm{sem}}
      +\alpha\mathcal{L}_{\mathrm{graph}},
    \label{eq:target_objective}
\end{equation}
where $\alpha$ balances the two loss terms. We initialize $\boldsymbol{c}=\boldsymbol{0}$ and optimize only the $r$ residual coefficients. The source detector, residual basis, target graph, and soft labels obtained from graph diffusion remain fixed throughout optimization. The learned coefficients determine how much the classifier is adjusted along each residual direction.





\subsection{Head Merging and Deployment}
\label{sec:deployment}

After residual optimization, we merge the learned correction into the source classifier. From Eq.~\eqref{eq:adapted_logit}, the adapted classifier weight and bias are
\begin{equation}
\begin{aligned}
    \boldsymbol{w}_t
    &=\boldsymbol{w}_s+\boldsymbol{U}\boldsymbol{c},\\
    b_t
    &=b_s-m_t
    -\boldsymbol{c}^{\top}\boldsymbol{U}^{\top}\bar{\boldsymbol{z}}.
\end{aligned}
\label{eq:final_head}
\end{equation}
The resulting classifier reproduces the adapted logits without additional score adjustment. During inference, only the frozen encoder and the updated linear head are used, with no additional inference branch. The algorithm of GraRA is summarized in the supplementary material (Sec.~A.1).

\begin{table*}[!thp]
\centering
\caption{Comparison with forensic detectors and unlabeled target
adaptation methods. Avg. is the unweighted macro average over
14 OpenMMSec~\cite{du2026sica} target domains.
\textbf{Bold} and \underline{underline} indicate the best and second-best results
in each column, respectively.}
\label{tab:sota_comparison}

\setlength{\tabcolsep}{6.5pt}
\renewcommand{\arraystretch}{0.8}
\small

\begin{tabular}{lll*{5}{cc}}
\toprule
\multirow{2}{*}{Group}
& \multirow{2}{*}{Method}
& \multirow{2}{*}{Venue}
& \multicolumn{2}{c}{Deepfake}
& \multicolumn{2}{c}{AIGC}
& \multicolumn{2}{c}{IMDL}
& \multicolumn{2}{c}{Document}
& \multicolumn{2}{c}{Avg.} \\
\cmidrule(lr){4-5}
\cmidrule(lr){6-7}
\cmidrule(lr){8-9}
\cmidrule(lr){10-11}
\cmidrule(lr){12-13}
& & & ACC & AUC & ACC & AUC & ACC & AUC
& ACC & AUC & ACC & AUC \\
\midrule

\multirow{7}{*}{Detector}
& UnivFD~\cite{ojha2023towards} & CVPR'23
& 75.97 & 85.99 & 74.53 & 86.43
& 75.03 & 84.18 & 55.12 & 56.66
& 72.24 & 81.73 \\

& AIDE~\cite{yan2025sanity} & ICLR'25
& 69.42 & 76.81 & 71.40 & 82.35
& 71.48 & 83.87 & 56.45 & 53.13
& 68.71 & 76.81 \\

& CO-SPY~\cite{cheng2025cospy} & CVPR'25
& 73.47 & 82.99 & 82.44 & 92.98
& 77.69 & 86.22 & 55.72 & 59.24
& 75.38 & 84.34 \\

& EFFORT~\cite{yan2025orthogonal} & ICML'25
& 87.21 & \underline{93.11} & 83.66 & \underline{93.84}
& \underline{79.69} & 89.00 & \underline{64.20} & 67.08
& 81.33 & 89.12 \\

& SICA~\cite{du2026sica} & ICML'26
& 86.06 & 90.17 & 82.41 & 91.88
& 78.24 & 88.19 & 62.57 & 70.71
& 80.03 & 87.84 \\

& SimLBR~\cite{dhakal2026simlbr} & CVPR'26
& 70.76 & 78.10 & 83.57 & 90.89
& 60.16 & 67.22 & 53.32 & 59.99
& 72.24 & 79.44 \\

& LTD~\cite{Yang_2026_CVPR} & CVPR'26
& 76.99 & 85.21 & 83.15 & 93.07
& 75.46 & 85.91 & 60.74 & 63.27
& 77.09 & 85.54 \\

\midrule

\multirow{5}{*}{Adaptation}
& TENT~\cite{wang2021tent} & ICLR'21
& 86.40 & 91.96 & 83.84 & 93.08
& 76.11 & 89.22 & 62.45 & 70.47
& 80.41 & 88.98 \\

& EATA~\cite{niu2022efficient} & ICML'22
& 86.00 & 91.65 & 83.03 & 92.87
& 78.31 & 88.17 & 62.21 & 70.70
& 80.23 & 88.69 \\

& SHOT~\cite{liang2020we} & ICML'20
& \underline{87.70} & 92.56 & \underline{86.93} & 93.52
& 78.63 & 88.25 & 57.64 & 68.16
& \underline{81.78} & 88.87 \\

& IST~\cite{ma2024improved} & CVPR'24
& 86.25 & 91.67 & 83.27 & 92.94
& 78.20 & 88.16 & 62.45 & 70.73
& 80.42 & 88.72 \\

& GOLD~\cite{lai2026golden} & CVPR'26
& 87.67 & \underline{93.11} & 83.79 & 93.54
& 78.37 & \underline{90.38} & 62.57 & \underline{71.42}
& 81.09 & \underline{89.81} \\

\midrule
\rowcolor{cvprblue!10}
Ours & \textbf{GraRA} & --
& \textbf{88.31} & \textbf{93.79}
& \textbf{88.33} & \textbf{95.17}
& \textbf{80.99} & \textbf{91.56}
& \textbf{67.92} & \textbf{74.77}
& \textbf{84.36} & \textbf{91.34} \\

\bottomrule
\end{tabular}
\vspace{-0.2cm}
\end{table*}

\section{Experiments}
\label{sec:experiments}
\subsection{Experimental Setup}
\label{sec:exp_setup}

\noindent\textbf{Datasets and protocol.}
We evaluate GraRA on OpenMMSec~\cite{du2026sica}, WildDeepfake~\cite{zi2020wilddeepfake}, and WildFake~\cite{hong2025wildfake}. The official OpenMMSec training split is used to train the source detector. Its validation split is divided into fixed, disjoint subsets for unlabeled adaptation and checkpoint and hyperparameter selection. The remaining 72 unseen forgery types form 14 target domains, covering six AIGC, four Deepfake, two image manipulation, and two document forgery domains. Each target domain is divided into unlabeled adaptation and held-out test sets at a 4:1 ratio. WildDeepfake and WildFake follow the same protocol. Further details are provided in the supplementary material (Sec.~A.2).

\noindent\textbf{Baselines and metrics.} We compare GraRA with seven fake image detectors: UnivFD~\cite{ojha2023towards}, AIDE~\cite{yan2025sanity}, CO-SPY~\cite{cheng2025cospy}, EFFORT~\cite{yan2025orthogonal}, SICA~\cite{du2026sica}, SimLBR~\cite{dhakal2026simlbr} and LTD~\cite{Yang_2026_CVPR}. We reproduce these detectors on the same OpenMMSec source data using their official procedures. We also compare five source-free adaptation methods: TENT~\cite{wang2021tent}, EATA~\cite{niu2022efficient}, SHOT~\cite{liang2020we}, IST~\cite{ma2024improved}, and GOLD~\cite{lai2026golden}, using their official implementations. All adaptation methods start from the same source checkpoint and use the same unlabeled target adaptation sets and held-out test sets as GraRA. We report ACC and AUC (\%) and their unweighted macro averages across target domains. 

\noindent\textbf{Implementation.}
Following ForensicHub~\cite{du2026forensichub}, we use its preprocessing and augmentation settings. The source detector uses CLIP ViT-L/14 with rank-8 LoRA ($\alpha_{\mathrm{LoRA}}=16$) applied to attention projections and a linear binary head. We train it for 10 epochs using AdamW, with a learning rate decaying from $10^{-4}$ to $10^{-5}$, weight decay 0.05, batch size 12, and gradient accumulation over eight steps. 

For target adaptation, we extract frozen features in mixed precision with batch size 32. We set the residual rank $r=32$, extra candidate dimension $\delta=8$, cosine $k$-NN graph size $k=240$, temperature $\tau=0.07$, and anchor selection ratio $\zeta=0.10$ for each score tail. We use $\lambda_{\mathrm{weak}}=0.05$, $\lambda_{\mathrm{anc}}=0.95$, and 50 diffusion steps. We initialize the residual coefficients to zero and optimize them using Adam for 300 full-batch steps over the entire unlabeled target adaptation set, with a learning rate of 0.02 and loss weight $\alpha=0.8$. All adaptation hyperparameters are kept fixed across all target domains. See supplementary material (Sec.~A.3) for key hyperparameter selection details.
\subsection{Comparison with SOTA Methods}
\label{sec:sota_comparison}

\noindent\textbf{Comparison with Forensic Detectors.}
As shown in Table~\ref{tab:sota_comparison}, GraRA achieves the highest macro-average ACC (84.36\%) and AUC (91.34\%). It outperforms EFFORT, the strongest forensic detector, by 3.03 and 2.22 points, respectively. These improvements show that adapting the source classifier with unlabeled target data can further improve detection performance over existing forensic detectors.

\noindent\textbf{Comparison with Target Adaptation Methods.}
GraRA also outperforms existing target adaptation methods. Among the adaptation baselines, SHOT achieves the highest ACC, while GOLD obtains the highest AUC. GraRA exceeds these two baselines by 2.58 and 1.53 percentage points in macro-average ACC and AUC, respectively. In addition, GraRA achieves the best ACC and AUC across all four categories, with consistent improvements on the challenging document domains. The AUC gains further indicate that GraRA improves the ranking of real and fake samples, rather than only adjusting the classification threshold. Per-domain results and plug-and-play adaptation to EFFORT are reported in the supplementary material (Sec.~C).

\begin{table}[!thp]
\centering
\caption{Component ablations and alternative designs on 14 OpenMMSec targets. Real and Fake denote classwise accuracies. Gap is the macro average of the per-domain absolute difference between real and fake ACC. Bold indicates the best result.}
\label{tab:ablation}
\setlength{\tabcolsep}{4.2pt}
\renewcommand{\arraystretch}{0.8}
\small
\begin{tabular}{@{}l*{5}{c}@{}}
\toprule
Variant & ACC & AUC & Real & Fake & Gap $\downarrow$ \\
\midrule
Source Only
& 80.03 & 87.84 & \textbf{88.16} & 71.90 & 18.63 \\
Shift Only
& 82.19 & 87.84 & 81.36 & 83.00 & 3.55 \\
\midrule
\rowcolor{gray!10}
\multicolumn{6}{c}{\textit{Effect of Individual Components}} \\
\cmidrule(lr){1-6}
w/o Orthogonalization
& 83.53 & 89.54 & 82.99 & 84.07 & 2.86 \\
w/o Semantic Anchors
& 83.27 & 88.46 & 82.72 & 83.81 & 3.00 \\
w/o Semantic Diffusion
& 83.27 & 88.30 & 82.66 & 83.87 & 3.04 \\
w/o Graph Consistency
& 82.82 & 88.18 & 82.07 & 83.56 & 3.45 \\
\midrule
\rowcolor{gray!10}
\multicolumn{6}{c}{\textit{Comparison with Alternative Designs}} \\
\cmidrule(lr){1-6}
Full Linear Head
& 80.05 & 88.13 & 82.07 & 78.03 & 4.38 \\
Random Orth. Basis
& 81.97 & 88.28 & 81.26 & 82.68 & 3.36 \\
Standard PCA
& 82.29 & 88.41 & 81.79 & 82.79 & 2.96 \\
Pseudo-label Training
& 81.62 & 87.71 & 81.35 & 81.88 & 2.60 \\
\midrule
\rowcolor{cvprblue!10}
\textbf{GraRA}
& \textbf{84.36} & \textbf{91.34} & 83.51
& \textbf{85.21} & \textbf{2.09} \\
\bottomrule
\end{tabular}
\vspace{-4mm}
\end{table}

\begin{figure*}[!thp]
    \centering
    \begin{minipage}[b]{0.33\textwidth}
        \centering
        \includegraphics[width=\linewidth]{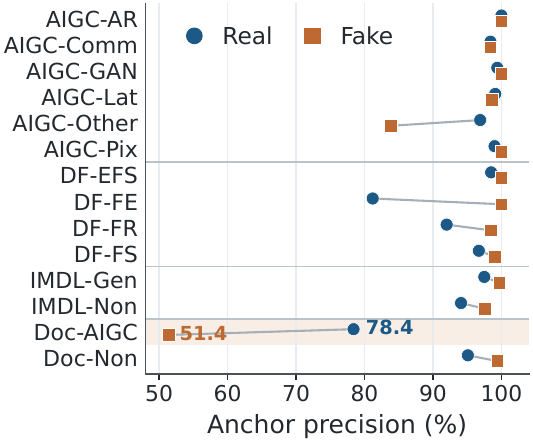}
        \par\vspace{2pt}
        {\footnotesize\textbf{(a) Anchor Quality}}
    \end{minipage}\hfill%
    \begin{minipage}[b]{0.33\textwidth}
        \centering
        \includegraphics[width=\linewidth]{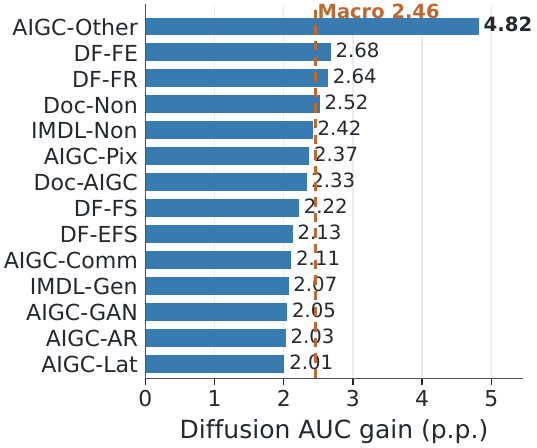}
        \par\vspace{2pt}
        {\footnotesize\textbf{(b) Diffusion Benefit}}
    \end{minipage}\hfill%
    \begin{minipage}[b]{0.33\textwidth}
        \centering
        \includegraphics[width=\linewidth]{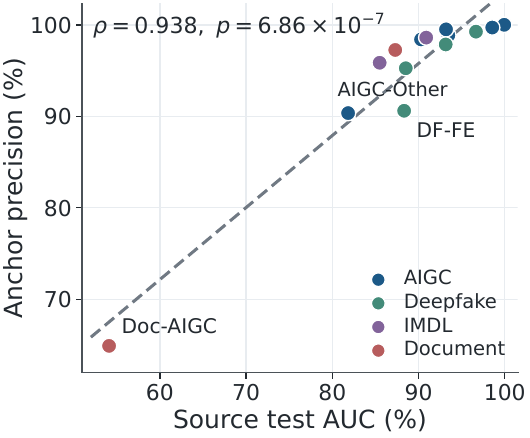}
        \par\vspace{2pt}
        {\footnotesize\textbf{(c) Source-Anchor Correlation}}
    \end{minipage}
    \vspace{-2mm}
    \caption{\textbf{Reliability of graph semantic guidance across 14 OpenMMSec targets.}
    (a) Precision of selected real and fake anchors.
    (b) AUC improvement of diffused soft labels over source predictions on the adaptation set.
    (c) Relationship between source test AUC and anchor precision, with Spearman $\rho$ and $p$-value.}
    \label{fig:semantic_reliability}
    \vspace{-4mm}
\end{figure*}

\subsection{Ablation Study}
\label{sec:component_mechanism}

We evaluate score shifting, component ablations, and alternative designs on 14 OpenMMSec targets under the same source checkpoint and target splits (Table~\ref{tab:ablation}). \emph{Source Only} refers to the pretrained source detector without adaptation. \emph{Shift Only} applies the reference-score shift without residual learning. \emph{w/o Semantic Anchors} replaces hard anchor labels with centered source probabilities $\widetilde{p}_i$. \emph{Full Linear Head} optimizes all 769 classifier parameters using the same soft labels and loss. \emph{Random Orthogonal Basis} uses 32 random orthogonal directions, while \emph{Standard PCA} removes uncertainty weighting. \emph{Pseudo-label Training} uses hard labels from centered source predictions without graph consistency.

\noindent\textbf{Effect of individual components.}
Shift Only improves ACC from 80.03\% to 82.19\% and reduces the class gap from 18.63 to 3.55 points, while AUC remains unchanged. This shows that score shifting improves class balance but cannot correct ranking errors. GraRA further increases ACC to 84.36\% and AUC to 91.34\%, while reducing the gap to 2.09 points. Removing any component reduces AUC, with the largest drop of 3.16 points observed without graph consistency. Removing semantic diffusion or hard anchor initialization also reduces AUC, supporting the benefit of graph-refined supervision. Removing orthogonalization reduces AUC and increases the class gap, supporting the use of source-orthogonal residual updates.

\noindent\textbf{Comparison with alternative designs.}
Full Linear Head updates 769 parameters but yields only marginal improvements over Source Only. In contrast, GraRA optimizes only 32 residual coefficients and outperforms Full Linear Head by 4.31 and 3.21 points in ACC and AUC, respectively. This suggests that constrained residual learning is more effective than updating the full classifier under noisy target supervision. An efficiency comparison is provided in the supplementary material (Sec.~E). Random Orthogonal Basis and Standard PCA underperform GraRA, supporting the use of uncertainty-weighted target features to construct the residual subspace. Pseudo-label Training performs worse than GraRA, showing the benefit of combining graph-refined soft labels with neighborhood consistency. These results support graph-guided supervision and constrained residual learning for source-free boundary adaptation.

\begin{figure*}[!thp]
    \centering
    \begin{minipage}[b]{0.32\textwidth}
        \centering
        \includegraphics[width=\linewidth]{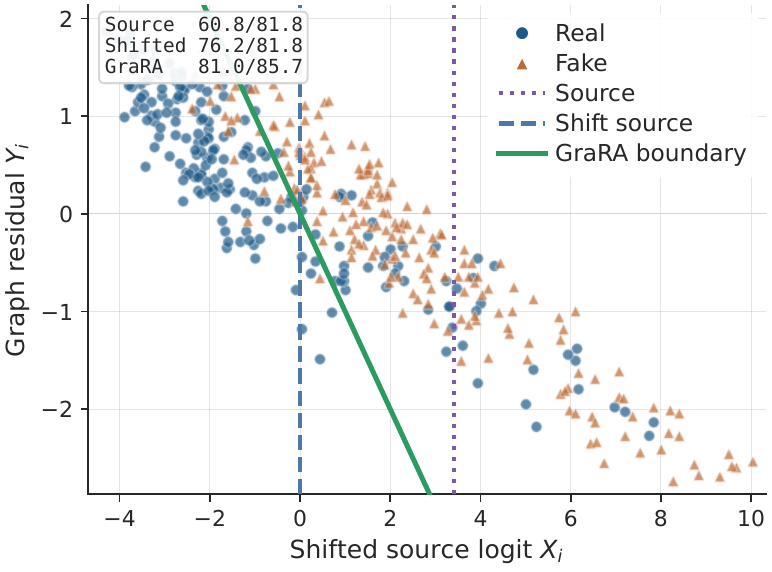}
        \par\vspace{2pt}
        {\footnotesize\textbf{(a) AIGC-Other Boundary}}
    \end{minipage}\hfill%
    \begin{minipage}[b]{0.32\textwidth}
        \centering
        \includegraphics[width=\linewidth]{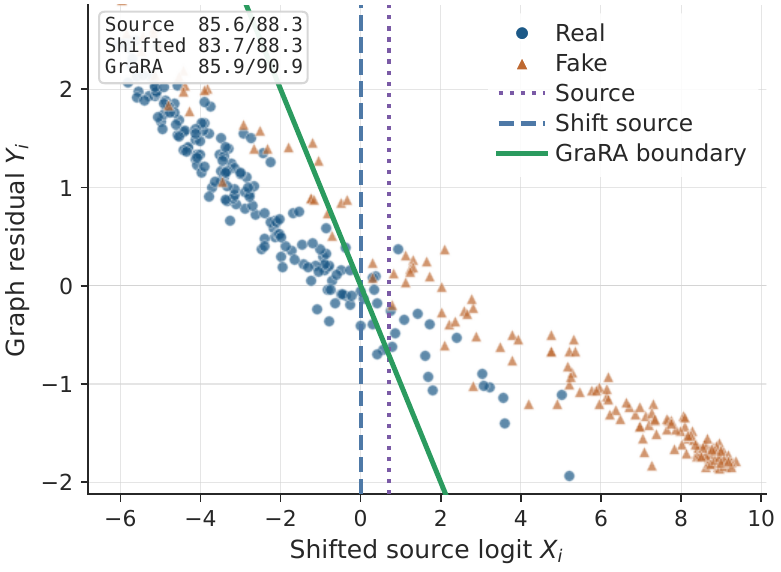}
        \par\vspace{2pt}
        {\footnotesize\textbf{(b) Deepfake-FE Boundary}}
    \end{minipage}\hfill%
    \begin{minipage}[b]{0.32\textwidth}
        \centering
        \includegraphics[width=\linewidth]{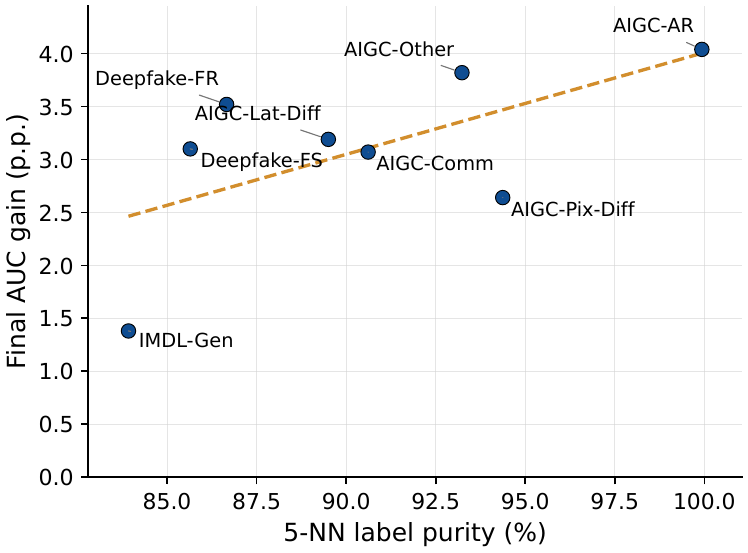}
        \par\vspace{2pt}
        {\footnotesize\textbf{(c) Local Neighborhood Structure}}
    \end{minipage}
    \vspace{-2mm}
    \caption{\textbf{Understanding boundary adaptation in GraRA.}
    (a, b) Decision boundaries of the source classifier, shifted source classifier, and GraRA on two representative targets.
    (c) Relationship between 5-NN label purity and final AUC gain on unseen target domains.}
    \label{fig:mechanism_analysis}
    \vspace{-2mm}
\end{figure*}

\subsection{Mechanism Analysis}
\label{sec:mechanism_analysis}

We analyze two aspects of GraRA: whether graph diffusion provides reliable guidance for residual learning, and how the learned residual changes the decision boundary.

\noindent\textbf{Reliability of Graph-Based Semantic Guidance.}
GraRA uses high-confidence source predictions as anchors and propagates their information over the target graph to generate soft labels. We evaluate the reliability of this guidance. Fig.~\ref{fig:semantic_reliability}(a, b) shows that selected anchors achieve 94.73\% mean precision, while graph diffusion improves average pseudo-label AUC by 2.46 points across 14 domains. These results show that graph diffusion spreads reliable anchor information to improve supervision for residual learning.

Anchor reliability is closely related to the discriminative ability of the source detector. As shown in Fig.~\ref{fig:semantic_reliability}(c), source AUC is positively correlated with anchor precision ($\rho=0.938$, $p=6.86\times10^{-7}$). This trend holds across most target domains, suggesting that a discriminative source detector generally provides more reliable anchors. One exception is Document-AIGC, where source AUC is close to chance level and fake-anchor precision drops sharply. Although diffusion still improves pseudo-label AUC in this domain, this case shows that reliable semantic guidance is difficult to obtain when source predictions provide weak class separation. Further analysis of challenging cases is provided in the supplementary material (Sec.~C.3).

\noindent\textbf{Understanding Decision Boundary Adaptation.}
We examine how GraRA changes the decision boundary. Fig.~\ref{fig:mechanism_analysis}(a, b) compares the source boundary, shifted boundary, and GraRA on two representative targets. A uniform logit shift changes the classification threshold without changing sample ranking. Thus, its effect on ACC varies across domains, while AUC remains unchanged. In contrast, GraRA improves ACC and AUC on both targets. This indicates that learned residuals can adjust the classification direction and correct ranking errors that a uniform shift cannot address.

We further examine the relationship between local feature structure and adaptation performance. Fig.~\ref{fig:mechanism_analysis}(c) shows a moderate positive correlation between 5-NN label purity and AUC gains. This suggests that locally consistent features may benefit boundary adaptation, although the relationship is not uniform across domains. Improving adaptation when local neighborhood structure is less reliable and exploring whether this relationship holds more generally remain promising directions for future work. Overall, these results show how GraRA uses graph structure to guide residual learning and adjust the source decision boundary.

\subsection{External Generalization}
\label{sec:generalization}

\begin{table}[!thp]
\centering
\caption{External generalization on WildDeepfake and WildFake. Source Only and EFFORT are evaluated without adaptation. Adaptation methods (TENT, GOLD) use unlabeled target data and are evaluated on held-out test sets. Avg. denotes the two-dataset macro average. Bold indicates the best result in each column.}
\label{tab:external_generalization}
\renewcommand{\arraystretch}{0.8}
\setlength{\tabcolsep}{5pt}
\small
\begin{tabular}{@{}lcccccc@{}}
\toprule
& \multicolumn{2}{c}{WildDeepfake}
& \multicolumn{2}{c}{WildFake}
& \multicolumn{2}{c}{Avg.} \\
\cmidrule(lr){2-3}
\cmidrule(lr){4-5}
\cmidrule(l){6-7}
Method & ACC & AUC & ACC & AUC & ACC & AUC \\
\midrule
Source Only
& 68.90 & 77.53 & 77.55 & 84.88
& 73.23 & 81.21 \\
EFFORT~\cite{yan2025orthogonal}
& 71.00 & 80.09 & 77.08 & 86.83
& 74.04 & 83.46 \\
\midrule
TENT~\cite{wang2021tent}
& 69.40 & 77.82 & 74.56 & 83.50
& 71.98 & 80.66 \\
GOLD~\cite{lai2026golden}
& 69.50 & 77.70 & 81.82 & \textbf{89.41}
& 75.66 & 83.56 \\
\midrule
\rowcolor{cvprblue!10}
\textbf{GraRA}
& \textbf{72.00} & \textbf{80.91}
& \textbf{82.17} & 88.81
& \textbf{77.09} & \textbf{84.86} \\
\bottomrule
\end{tabular}
\vspace{-0.4cm}
\end{table}

To evaluate GraRA beyond OpenMMSec, we conduct unlabeled target adaptation on WildDeepfake and WildFake using fixed hyperparameters. As shown in Table~\ref{tab:external_generalization}, GraRA improves both ACC and AUC over Source Only on both datasets. It also achieves the highest macro-average ACC and AUC, outperforming GOLD~\cite{lai2026golden}, the strongest adaptation baseline in this comparison. Although GOLD achieves a slightly higher AUC on WildFake, GraRA performs better overall across the two datasets. These results show that GraRA remains effective on external forensic domains. Additional analyses of target class-ratio sensitivity and robustness to perturbations are provided in the supplementary material (Secs.~D.1--D.2).

\section{Conclusion}
\label{sec:conclusion}

We revisit cross-domain fake image detection from the perspective of decision boundary mismatch. Our diagnostic experiments show that frozen forensic features preserve discriminative information that the source classifier does not fully exploit. We propose GraRA to adapt the source decision boundary using only unlabeled target images. GraRA constrains classifier updates to an uncertainty-aware residual subspace orthogonal to the source classification direction, while graph-diffused soft labels and neighborhood consistency jointly guide residual learning. Extensive experiments on unseen domains demonstrate improvements in both ACC and AUC, while ablation and mechanism analyses support its effectiveness. These findings suggest that decision boundary adaptation can complement representation learning for cross-domain fake image detection.

\section*{LLM Usage Statement}
In this study, OpenAI's ChatGPT was used to assist with language polishing, figure optimization, and project code organization. Its use mainly focused on improving the clarity of writing, the presentation of figures, and the structure of the code.

{
    \small
    \bibliographystyle{ieeenat_fullname}
    \bibliography{main}
}

\appendix

\clearpage
\setcounter{page}{1}
\maketitlesupplementary

\section*{Overview}

This supplementary material provides additional details
on GraRA and its experimental evaluation. It is organized
as follows:

\begin{itemize}
    \item \textbf{Sec.~\ref{app:method_details}}
    presents the complete algorithm, dataset and evaluation protocols,
    and source-validation hyperparameter selection.

    \item \textbf{Sec.~\ref{app:diagnostics}}
    describes the diagnostic experiments, including
    target-supervised linear probing, local
    neighborhood purity, graph quality and semantic guidance analysis.

    \item \textbf{Sec.~\ref{app:detailed_eval}}
    provides detailed cross-domain results on OpenMMSec, evaluates the adaptation of EFFORT using GraRA, analyzes challenging cases and future directions.

    \item \textbf{Sec.~\ref{app:sensitivity_robustness}}
    examines sensitivity to target class ratios and robustness to image perturbations.

    \item \textbf{Sec.~\ref{sec:efficiency}}
    reports computational efficiency in terms of
    optimized parameters, adaptation time, memory
    usage, and inference latency.
\end{itemize}

\begin{algorithm}[!thp]
\caption{GraRA}
\label{alg:GraRA}
\small
\begin{algorithmic}[1]
\Require Unlabeled target images
$D_t^u=\{\boldsymbol{x}_i\}_{i=1}^{N_t}$,
frozen encoder $E_{\boldsymbol{\theta}_s}$,
source head $(\boldsymbol{w}_s,b_s)$
\Ensure Adapted linear head $(\boldsymbol{w}_t,b_t)$

\Statex \textbf{Step 1: Feature Extraction and Score Reference}
\State $\boldsymbol{z}_i \gets E_{\boldsymbol{\theta}_s}(\boldsymbol{x}_i)$
\State $s_i \gets \boldsymbol{w}_s^\top\boldsymbol{z}_i+b_s$
\State $\mathcal{A}_{\mathrm r}
\gets \operatorname{Bottom}_{\zeta}(\{s_i\})$
\State $\mathcal{A}_{\mathrm f}
\gets \operatorname{Top}_{\zeta}(\{s_i\})$
\State $m_{\mathrm r}\gets
\operatorname{median}_{i\in\mathcal{A}_{\mathrm r}}s_i$
\State $m_{\mathrm f}\gets
\operatorname{median}_{i\in\mathcal{A}_{\mathrm f}}s_i$
\State $m_t\gets(m_{\mathrm r}+m_{\mathrm f})/2$
\State $\widetilde{s}_i\gets s_i-m_t$,
$\widetilde{p}_i\gets\sigma(\widetilde{s}_i)$

\Statex \textbf{Step 2: Residual Subspace Construction}
\State $q_i\gets
\max\{\epsilon,4\widetilde{p}_i(1-\widetilde{p}_i)\}$
\State $\boldsymbol{C}_q\gets
\operatorname{WeightedCov}(\{\boldsymbol{z}_i,q_i\})$
\State $\boldsymbol{A}_{r+\delta}\gets
\operatorname{TopEig}_{r+\delta}(\boldsymbol{C}_q)$
\State $\bar{\boldsymbol{w}}_s\gets
\boldsymbol{w}_s/\|\boldsymbol{w}_s\|_2$
\State $\boldsymbol{P}_{\perp}\gets
\boldsymbol{I}-
\bar{\boldsymbol{w}}_s\bar{\boldsymbol{w}}_s^\top$
\State $\boldsymbol{U}\gets
\operatorname{orth}_r(
\boldsymbol{P}_{\perp}\boldsymbol{A}_{r+\delta})$

\Statex \textbf{Step 3: Source-Guided Graph Diffusion}
\State Build a cosine $k$NN graph with affinities $a_{ij}$
\State Set $g_i^{(0)}\gets 0$ for $i\in\mathcal{A}_{\mathrm r}$
\State Set $g_i^{(0)}\gets 1$ for $i\in\mathcal{A}_{\mathrm f}$
\State Set $g_i^{(0)}\gets\widetilde{p}_i$ otherwise
\State Compute confidence weights $\lambda_i$
using Eq.~\eqref{eq:supp_confidence_tether}
\State Set $\lambda_i\gets\lambda_{\mathrm{anc}}$
for anchor samples
\For{$t=0,\ldots,T-1$}
    \State Update all $g_i^{(t+1)}$ using
    Eq.~\eqref{eq:supp_diffusion}
\EndFor
\State $\bar{g}_i\gets g_i^{(T)}$

\Statex \textbf{Step 4: Graph-Consistent Residual Learning}
\State $\bar{\boldsymbol{z}}\gets
\frac{1}{N_t}\sum_i\boldsymbol{z}_i$
\State $\boldsymbol{h}_i\gets
\boldsymbol{U}^\top(
\boldsymbol{z}_i-\bar{\boldsymbol{z}})$
\State Initialize $\boldsymbol{c}\gets\boldsymbol{0}$
\For{$u=1,\ldots,K$}
    \State $\ell_i\gets
    \widetilde{s}_i+\boldsymbol{c}^\top\boldsymbol{h}_i$
    \State Compute $\mathcal{L}_{\mathrm{sem}}$
    and $\mathcal{L}_{\mathrm{graph}}$
    \State $\mathcal{L}\gets
    (1-\alpha)\mathcal{L}_{\mathrm{sem}}
    +\alpha\mathcal{L}_{\mathrm{graph}}$
    \State Update $\boldsymbol{c}$ using
    $\nabla_{\boldsymbol{c}}\mathcal{L}$
\EndFor

\Statex \textbf{Step 5: Head Merging}
\State $\boldsymbol{w}_t\gets
\boldsymbol{w}_s+\boldsymbol{U}\boldsymbol{c}$
\State $b_t\gets
b_s-m_t-
\boldsymbol{c}^\top\boldsymbol{U}^\top
\bar{\boldsymbol{z}}$
\State \Return $(\boldsymbol{w}_t,b_t)$

\end{algorithmic}
\end{algorithm}

\section{Method and Implementation Details}
\label{app:method_details}

\subsection{Algorithm of GraRA}
\label{app:algorithm}

Algorithm~\ref{alg:GraRA} summarizes GraRA. The frozen detector extracts
features and source logits from unlabeled target images. A score reference
is estimated from the two prediction tails, after which uncertainty-aware
PCA restricts classifier updates to a source-orthogonal residual subspace.
Source-guided diffusion produces soft targets on a feature graph, and only
the residual coefficients are optimized. The resulting correction is
merged into the source head without an additional inference branch.

For completeness, the confidence weight of each non-anchor sample and
the graph diffusion update are defined here:
\begin{equation}
\lambda_i=\lambda_{\mathrm{weak}}
 +(\lambda_{\mathrm{anc}}-\lambda_{\mathrm{weak}})
 |2\widetilde{p}_i-1|,
\label{eq:supp_confidence_tether}
\end{equation}
\begin{equation}
g_i^{(t+1)}=\lambda_i g_i^{(0)}
 +(1-\lambda_i)\sum_{j\in\mathcal{N}_k(i)}a_{ij}g_j^{(t)}.
\label{eq:supp_diffusion}
\end{equation}
Here $\lambda_i=\lambda_{\mathrm{anc}}$ for anchors,
$\mathcal{N}_k(i)$ denotes the cosine-$k$NN neighborhood, and
$a_{ij}$ denotes the row-normalized similarity weight.
The final loss combines soft-target cross-entropy and graph-weighted
neighborhood consistency.

\begin{figure*}[!thp]
\centering
\includegraphics[width=\textwidth]{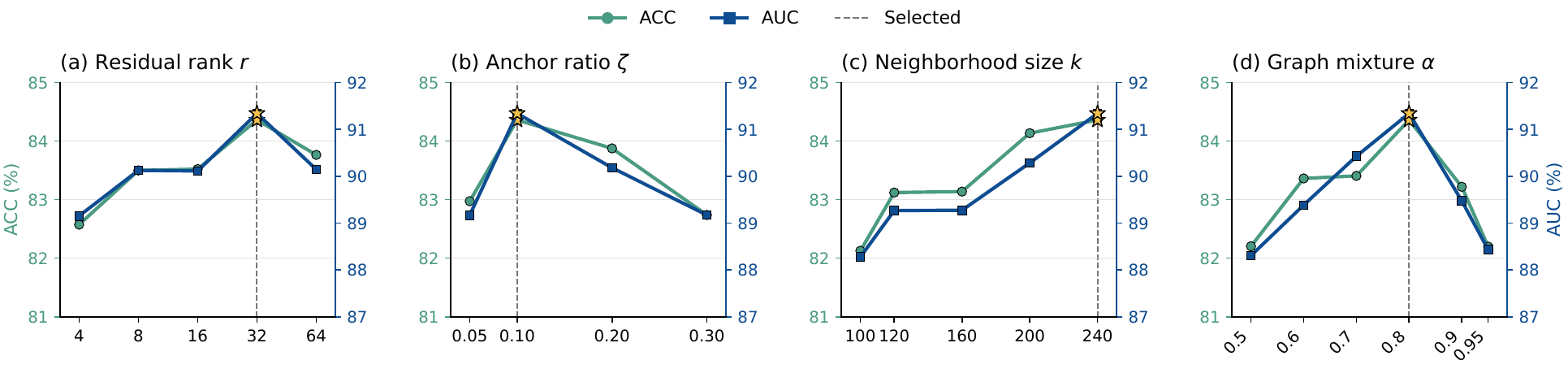}
\caption{Hyperparameter selection on the OpenMMSec source validation split. We vary (a) residual rank $r$, (b) anchor selection ratio $\zeta$, (c) neighborhood size $k$, and (d) graph loss weight $\alpha$, while keeping other settings fixed. ACC and AUC are evaluated on the selection subset, which is disjoint from the adaptation subset. Dashed vertical lines indicate the selected values used in the main experiments.}
\label{fig:hyperparameter_selection}
\end{figure*}

\subsection{Dataset and Evaluation Protocols}
\label{app:dataset_details}

\subsubsection{Datasets}

\noindent\textbf{OpenMMSec.}
OpenMMSec~\cite{du2026sica} covers four forensic tasks: Deepfake, AI-generated image (AIGC), image manipulation (IMDL), and document forgery detection. Following its official protocol, 26 forgery types are designated for source training and validation, while the remaining 72 unseen types form 14 target domains. These include four Deepfake domains (EFS, FE, FR, FS), six AIGC domains (GAN, latent diffusion, pixel diffusion, autoregressive, commercial, and others), two IMDL domains (generative and non-generative), and two document forgery domains (AIGC and non-AIGC).

\noindent\textbf{WildDeepfake.}
WildDeepfake~\cite{zi2020wilddeepfake} is a real-world Deepfake dataset collected from the Internet. It contains 7,314 face sequences extracted from 707 videos, covering diverse scenes, facial appearances, and image qualities. Its diverse online sources make it useful for evaluating adaptation to real-world Deepfake images.

\noindent\textbf{WildFake.}
WildFake~\cite{hong2025wildfake} is a large-scale AI-generated image detection dataset containing 2,557,278 fake images and 1,013,446 real images. Its fake images cover GANs, diffusion models, and other generative architectures, with variations in generator types, model weights, and versions. This diversity provides an additional benchmark for evaluating adaptation to unseen AI-generated images.

\subsubsection{Evaluation Protocols}

\noindent\textbf{Source Detector Training and Checkpoint Selection.}
We train the source detector on the official OpenMMSec training split. The source validation split is divided once into two fixed, disjoint subsets using a stratified 4:1 split: an adaptation subset and a selection subset. We select the source checkpoint with the highest AUC on the selection subset. This checkpoint is then fixed for all subsequent experiments.

\noindent\textbf{Hyperparameter Selection.}
For hyperparameter selection, GraRA uses only unlabeled images from the source-validation adaptation subset. The selection subset is excluded from all adaptation operations, including feature graph construction, anchor selection, and residual optimization. Its labels are used to evaluate the adapted classifiers using ACC and AUC. We use the same fixed partition for all candidate configurations and select the four hyperparameters independently, as detailed in Sec.~\ref{app:hyperparameter_selection}.

\noindent\textbf{OpenMMSec Target Evaluation.}
For each of the 14 target domains, we randomly partition all samples into two disjoint subsets: 80\% for unlabeled adaptation and 20\% for testing. All adaptation methods use the same unlabeled adaptation subset, and all compared methods are evaluated on the same held-out test subset. Target labels are not used for adaptation or hyperparameter selection.

\noindent\textbf{External Target Evaluation.}
We use the official test sets of WildDeepfake and WildFake as two external target domains. Each dataset is independently divided into unlabeled adaptation and held-out test subsets at a 4:1 ratio. For WildDeepfake, the split is performed at the video level to ensure that frames from the same video do not appear in both subsets. GraRA uses the source checkpoint and hyperparameters selected on OpenMMSec without further adjustment. The encoder remains frozen, and only the residual coefficients are optimized. Source Only and EFFORT are evaluated without adaptation, while TENT~\cite{wang2021tent} and GOLD~\cite{lai2026golden} use the same source checkpoint and target partitions as GraRA. No target labels are used for adaptation or hyperparameter selection, and test samples are excluded from adaptation.

\subsection{Hyperparameter Selection}
\label{app:hyperparameter_selection}

We select the four hyperparameters using the fixed partition of the OpenMMSec source validation split described in Sec.~\ref{app:dataset_details}. Specifically, the validation data are divided into two disjoint subsets: an adaptation subset and a selection subset. For each candidate configuration, GraRA is applied using only unlabeled images from the adaptation subset. The selection subset is excluded from all adaptation steps, including graph construction, anchor selection, and residual optimization. After adaptation, the resulting classifier is evaluated on the selection subset, using its labels to compute ACC and AUC. The same partition is used for all candidate configurations, and we vary one hyperparameter at a time while keeping the others fixed.

As shown in Fig.~\ref{fig:hyperparameter_selection}, $r=32$ achieves the best performance, while performance declines at $r=64$. Among the tested values, $\zeta=0.10$ and $\alpha=0.8$ yield the best results. For neighborhood size, we evaluate $k\in\{100,120,160,200,240\}$. Both metrics improve with larger neighborhoods within the tested range. We select $k=240$, which provides broader neighborhood coverage while maintaining a sparse graph. The selected hyperparameters are fixed for all subsequent target-domain experiments. No target-domain data or labels are used for hyperparameter selection, and source validation images are not accessed during target-time adaptation.

\section{Details of Diagnostic Experiments}
\label{app:diagnostics}

\subsection{Target-Supervised Linear Probe}
\label{app:diagnostic_probe}

For each of the 14 OpenMMSec target domains, we use the same 80/20 target split as GraRA: 80\% of the samples for probe training and the remaining 20\% for independent testing. These correspond to the same adaptation and test sets used by GraRA. We freeze the source-trained feature extractor described in the implementation details and extract features from both subsets separately. A linear logistic regression classifier is trained using only labeled target training features. Feature standardization and classification are implemented in a single pipeline. The regularization parameter is selected from $\{0.01, 0.1, 1, 10, 100\}$ using five-fold stratified cross-validation on the training set. The selected pipeline is then refitted on the full target training set and evaluated once on the held-out target test set. No test labels are used for training or hyperparameter selection. Across the 14 domains, the probe achieves a macro-average ACC of 93.76\% and AUC of 98.02\%. We emphasize that this target-supervised probe serves only as a retrospective target-supervised diagnostic reference. It measures the class information that remains linearly accessible in the frozen features and does not represent the performance achievable by our unlabeled adaptation method.

\subsection{Local Neighborhood Purity}
\label{app:diagnostic_purity}

We also measure label purity in the frozen target test feature space. Features are $\ell_2$-normalized, and nearest neighbors are identified using cosine distance. The query sample itself is excluded by retrieving $k+1$ neighbors and removing the zero-distance self-match. The reported 5-NN purity is the average fraction of the remaining five neighbors that share the query label. We use $k=5$ as a representative small-neighborhood measure, rather than a tuned hyperparameter. A single neighbor ($k=1$) is more sensitive to individual samples, while larger neighborhoods gradually include less local structure. The results remain consistent for $k\in\{1,5,10,20\}$, with macro-average purity values of 92.10\%, 90.82\%, 90.16\%, and 89.51\%, respectively. Target labels are used only to compute this retrospective metric and are not provided to GraRA.

\subsection{Graph Quality and Semantic Guidance}
\label{app:neighborhood_quality}

\begin{table}[!thp]
\centering
\caption{Graph quality and pseudo-label AUC (\%) across neighborhood sizes on 14 OpenMMSec adaptation sets. Results are macro-averaged over domains.}
\label{tab:graph_quality_knn}
\renewcommand{\arraystretch}{0.8}
\setlength{\tabcolsep}{16pt}
\small
\begin{tabular}{ccc}
\toprule
$k$ & Weighted purity & Pseudo-label AUC \\
\midrule
100 & 85.69 & 91.26 \\
120 & 85.39 & 91.26 \\
160 & 84.91 & 91.26 \\
200 & 84.53 & 91.26 \\
240 & 84.23 & 91.26 \\
\bottomrule
\end{tabular}
\end{table}

We further examine how neighborhood size affects graph quality and semantic diffusion. We vary $k\in\{100,120,160,200,240\}$ while keeping other settings fixed. Weighted label purity measures the normalized graph weight assigned to same-class neighbors, excluding self-links. Pseudo-label AUC evaluates the soft labels after diffusion. Both metrics are macro-averaged over 14 OpenMMSec adaptation sets, with target labels used only for retrospective evaluation.

As shown in Table~\ref{tab:graph_quality_knn}, weighted purity decreases from 85.69\% to 84.23\% as $k$ increases, while pseudo-label AUC remains stable at around 91.26\%. This low sensitivity may be explained by two factors. First, larger neighborhoods retain highly similar neighbors, while similarity weights limit the influence of additional neighbors. Second, fixed source predictions and semantic anchors provide consistent guidance across different values of $k$. Since AUC measures sample ranking, it can remain stable even when the diffusion probabilities change.

\section{Detailed Cross-Domain Evaluation}
\label{app:detailed_eval}

\subsection{Per-Domain Comparison on OpenMMSec}
\label{app:per_domain}

\begin{table*}[!thp]
\centering
\caption{Per-domain ACC (\%) on the 14 OpenMMSec targets. Avg. is the unweighted macro average across domains. Bold indicates the best result in each column, including ties.}
\label{tab:openmmsec_acc}
\setlength{\tabcolsep}{2.2pt}
\resizebox{\textwidth}{!}{%
\begin{tabular}{ll*{15}{c}}
\toprule
\multirow{2}{*}{Group}
& \multirow{2}{*}{Method}
& \multicolumn{4}{c}{Deepfake}
& \multicolumn{6}{c}{AIGC}
& \multicolumn{2}{c}{IMDL}
& \multicolumn{2}{c}{Document}
& \multirow{2}{*}{Avg.} \\
\cmidrule(lr){3-6}
\cmidrule(lr){7-12}
\cmidrule(lr){13-14}
\cmidrule(lr){15-16}
& & EFS & FE & FR & FS
& GAN & Lat. & Pix. & AR & Comm. & Other
& Gen. & Non-Gen.
& AIGC & Non-AIGC & \\
\midrule

\multirow{7}{*}{Detector}
& UnivFD~\cite{ojha2023towards}
& 80.31 & 82.87 & 70.38 & 70.33
& 86.14 & 63.20 & 70.42 & 98.75 & 59.17 & 69.50
& 71.46 & 78.61 & 38.04 & 72.20 & 72.24 \\

& AIDE~\cite{yan2025sanity}
& 75.26 & 75.14 & 67.12 & 60.17
& 68.86 & 63.56 & 77.92 & 95.00 & 60.83 & 62.25
& 74.63 & 68.33 & 40.22 & 72.68 & 68.71 \\

& CO-SPY~\cite{cheng2025cospy}
& 74.85 & 80.39 & 68.48 & 70.17
& 90.23 & 76.40 & 84.17 & \textbf{100.00} & 67.08 & 76.75
& 75.37 & \textbf{80.00} & 40.22 & 71.22 & 75.38 \\

& EFFORT~\cite{yan2025orthogonal}
& 90.10 & \textbf{91.71} & 82.20 & 84.83
& \textbf{96.59} & 81.02 & \textbf{87.50} & 98.75 & 79.38 & 58.75
& 80.77 & 78.61 & 51.09 & \textbf{77.32} & 81.33 \\

& SICA~\cite{du2026sica}
& 90.52 & 85.64 & 81.59 & 86.50
& 93.41 & 83.27 & 83.33 & 93.50 & 80.21 & 60.75
& 82.73 & 73.75 & 47.83 & \textbf{77.32} & 80.03 \\

& SimLBR~\cite{dhakal2026simlbr}
& 82.16 & 78.18 & 66.29 & 56.42
& 87.27 & 77.68 & 80.83 & 92.50 & 78.12 & \textbf{85.00}
& 58.37 & 61.94 & 43.48 & 63.17 & 72.24 \\

& LTD~\cite{Yang_2026_CVPR}
& 87.11 & 79.01 & 76.59 & 65.25
& 93.41 & 83.94 & 85.42 & 98.75 & 76.88 & 60.50
& 80.09 & 70.83 & 47.83 & 73.66 & 77.09 \\

\midrule

\multirow{5}{*}{Adaptation}
& TENT~\cite{wang2021tent}
& 91.96 & 85.64 & 80.91 & 87.08
& 93.64 & \textbf{86.56} & 84.58 & 97.50 & 82.29 & 58.50
& 78.74 & 73.47 & 47.83 & 77.07 & 80.41 \\

& EATA~\cite{niu2022efficient}
& 90.62 & 85.36 & 81.52 & 86.50
& 93.64 & 83.15 & 82.92 & 97.50 & 80.21 & 60.75
& 82.73 & 73.89 & 47.83 & 76.59 & 80.23 \\

& SHOT~\cite{liang2020we}
& 91.37 & 85.83 & 84.77 & 88.83
& 95.68 & 84.29 & 83.25 & 98.00 & 83.12 & 77.25
& 82.19 & 75.07 & 40.65 & 74.63 & 81.78 \\

& IST~\cite{ma2024improved}
& 90.93 & 85.64 & 81.52 & 86.92
& 93.64 & 84.00 & 83.33 & 97.50 & 80.63 & 60.50
& 82.93 & 73.47 & 47.83 & 77.07 & 80.42 \\

& GOLD~\cite{lai2026golden}
& 93.09 & 87.29 & 82.05 & 88.25
& 93.86 & 84.98 & 85.42 & 98.75 & 80.00 & 59.75
& 81.04 & 75.69 & 47.83 & \textbf{77.32} & 81.09 \\

\midrule
\rowcolor{cvprblue!10}
\textbf{Ours} & \textbf{GraRA}
& \textbf{93.24} & 85.91 & \textbf{85.18} & \textbf{88.92}
& 96.09 & 85.46 & 86.25 & 97.50 & \textbf{83.67} & 81.00
& \textbf{83.46} & 78.52 & \textbf{58.52} & \textbf{77.32}
& \textbf{84.36} \\

\bottomrule
\end{tabular}}
\end{table*}

\begin{table*}[!thp]
\centering
\caption{Per-domain AUC (\%) on the 14 OpenMMSec targets. Avg. is the unweighted macro average across domains. Bold indicates the best result in each column, including ties.}
\label{tab:openmmsec_auc}
\setlength{\tabcolsep}{2.2pt}
\resizebox{\textwidth}{!}{%
\begin{tabular}{ll*{15}{c}}
\toprule
\multirow{2}{*}{Group}
& \multirow{2}{*}{Method}
& \multicolumn{4}{c}{Deepfake}
& \multicolumn{6}{c}{AIGC}
& \multicolumn{2}{c}{IMDL}
& \multicolumn{2}{c}{Document}
& \multirow{2}{*}{Avg.} \\
\cmidrule(lr){3-6}
\cmidrule(lr){7-12}
\cmidrule(lr){13-14}
\cmidrule(lr){15-16}
& & EFS & FE & FR & FS
& GAN & Lat. & Pix. & AR & Comm. & Other
& Gen. & Non-Gen.
& AIGC & Non-AIGC & \\
\midrule

\multirow{7}{*}{Detector}
& UnivFD~\cite{ojha2023towards}
& 91.48 & \textbf{97.08} & 76.68 & 78.73
& 93.59 & 77.45 & 85.30 & \textbf{100.00} & 74.81 & 87.41
& 82.97 & 85.39 & 35.99 & 77.32 & 81.73 \\

& AIDE~\cite{yan2025sanity}
& 85.25 & 83.00 & 72.65 & 66.36
& 80.93 & 74.72 & 88.65 & 98.75 & 71.11 & 79.97
& 86.58 & 81.15 & 30.10 & 76.16 & 76.81 \\

& CO-SPY~\cite{cheng2025cospy}
& 85.41 & 94.12 & 75.76 & 76.67
& 97.37 & 89.25 & 95.69 & \textbf{100.00} & 85.96 & 89.59
& 86.24 & 86.19 & 39.51 & 78.97 & 84.34 \\

& EFFORT~\cite{yan2025orthogonal}
& 95.91 & 94.70 & 90.25 & 91.58
& 98.92 & 94.32 & \textbf{96.22} & \textbf{100.00} & 92.94 & 80.64
& 89.75 & 88.25 & 48.16 & 86.00 & 89.12 \\

& SICA~\cite{du2026sica}
& 93.66 & 88.33 & 88.52 & 90.15
& 96.55 & 93.46 & 93.19 & 95.94 & 90.29 & 81.83
& 90.89 & 85.49 & 54.11 & 87.30 & 87.84 \\

& SimLBR~\cite{dhakal2026simlbr}
& 90.32 & 87.96 & 72.67 & 61.44
& 94.72 & 84.60 & 88.47 & \textbf{100.00} & 84.44 & \textbf{93.09}
& 66.21 & 68.22 & 50.43 & 69.55 & 79.44 \\

& LTD~\cite{Yang_2026_CVPR}
& 96.17 & 88.98 & 83.98 & 71.69
& 98.21 & 94.47 & 93.26 & \textbf{100.00} & 89.23 & 83.28
& 90.51 & 81.31 & 43.27 & 83.27 & 85.54 \\

\midrule

\multirow{5}{*}{Adaptation}
& TENT~\cite{wang2021tent}
& 96.64 & 88.55 & 89.19 & 93.46
& 98.58 & 94.09 & 93.40 & 99.94 & 91.01 & 81.50
& 90.15 & 88.29 & 53.36 & 87.59 & 88.98 \\

& EATA~\cite{niu2022efficient}
& 96.65 & 88.30 & 88.53 & 93.14
& 98.55 & 93.46 & 93.17 & 99.94 & 90.29 & 81.81
& 90.89 & 85.46 & 54.11 & 87.30 & 88.69 \\

& SHOT~\cite{liang2020we}
& 97.03 & 89.29 & 90.76 & 93.16
& 99.15 & 93.71 & 95.53 & 99.90 & 92.03 & 80.81
& 89.78 & 86.72 & 55.47 & 80.84 & 88.87 \\

& IST~\cite{ma2024improved}
& 96.58 & 88.33 & 88.65 & 93.10
& 98.55 & 93.60 & 93.18 & 99.94 & 90.36 & 82.00
& 90.92 & 85.40 & 54.11 & 87.35 & 88.72 \\

& GOLD~\cite{lai2026golden}
& 96.85 & 90.51 & 90.90 & \textbf{94.19}
& 98.69 & 95.18 & 94.67 & 99.94 & 91.57 & 81.20
& 90.31 & 90.44 & 54.02 & \textbf{88.82} & 89.81 \\

\midrule
\rowcolor{cvprblue!10}
\textbf{Ours} & \textbf{GraRA}
& \textbf{98.92} & 90.93 & \textbf{92.04} & 93.25
& \textbf{99.55} & \textbf{96.65} & 95.83 & 99.98 & \textbf{93.36} & 85.65
& \textbf{92.27} & \textbf{90.84} & \textbf{60.88} & 88.65
& \textbf{91.34} \\

\bottomrule
\end{tabular}}
\end{table*}

Tables~\ref{tab:openmmsec_acc} and~\ref{tab:openmmsec_auc}
report ACC and AUC on all 14 OpenMMSec targets.
GraRA achieves the highest macro-average ACC and AUC among the
compared methods. The largest gains are +20.25 ACC points on AIGC-Other and
+6.77 AUC points on Document-AIGC. Despite this improvement, Document-AIGC remains challenging, suggesting limitations in boundary adaptation when source predictions provide weak class separation. Further analysis is provided in Sec.~\ref{app:challenging_cases}.

\begin{figure}[!thp]
\centering
\includegraphics[width=\linewidth]{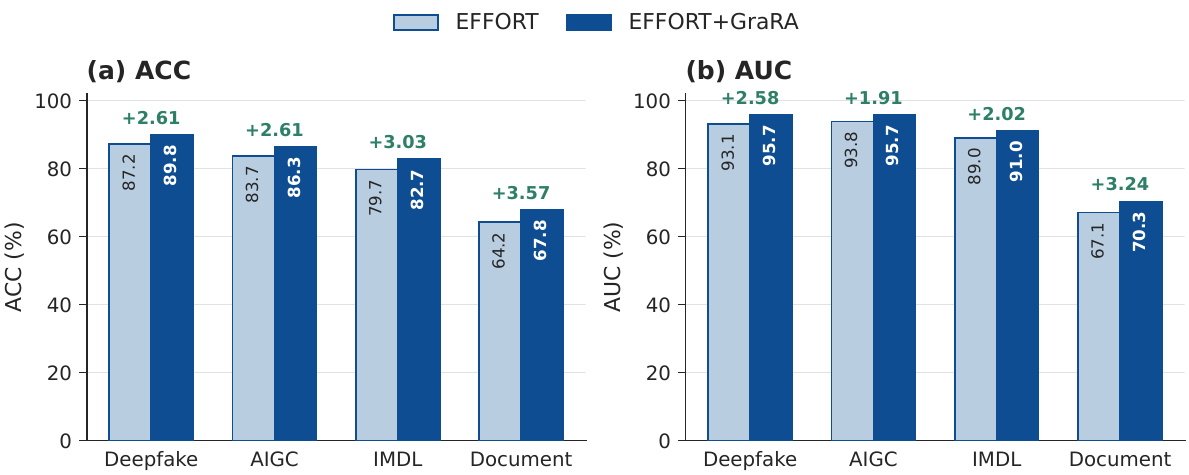}
\caption{Plug-and-play adaptation of EFFORT across four forgery categories on OpenMMSec. (a) ACC and (b) AUC before and after applying GraRA. Values are averaged within each category, and the numbers above the bars indicate improvements in percentage points. The EFFORT encoder remains frozen during adaptation.}
\label{fig:effort_dcgra}
\vspace{-0.3cm}
\end{figure}

\subsection{Plug-and-Play Adaptation to EFFORT}
\label{app:effort_adaptation}

To examine whether GraRA can improve other forensic detectors, we apply it to EFFORT, the strongest detector baseline in our main comparison. We freeze the EFFORT encoder and adapt its linear classification head using unlabeled target features. Figure~\ref{fig:effort_dcgra} compares EFFORT and EFFORT+GraRA across four forgery categories. GraRA improves both ACC and AUC in all categories, with gains of 2.61--3.57 and 1.91--3.24 percentage points, respectively. The largest gains occur on document forgery, where EFFORT shows relatively low performance. These results show that GraRA can also improve another pretrained detector without updating its feature encoder.

\subsection{Challenging Cases and Future Directions}
\label{app:challenging_cases}

\begin{table}[t]
\centering
\caption{Document-AIGC diagnostics (\%). Probe and 5-NN metrics are evaluated on the held-out target set; anchor and diffusion metrics are evaluated on the adaptation set. The target-supervised probe is diagnostic only.}
\label{tab:doc_aigc_diagnostics}
\setlength{\tabcolsep}{15pt}
\renewcommand{\arraystretch}{1}
\small
\begin{tabular}{lc}
\toprule
Metric & Value \\
\midrule
Source test ACC / AUC & 47.83 / 54.11 \\
Target-supervised probe ACC & 87.51 \\
5-NN label purity & 83.51 \\
Real / Fake anchor precision & 78.4 / 51.4 \\
Diffusion pseudo-label AUC gain & +2.33 p.p. \\
GraRA test ACC / AUC & 58.52 / 60.88 \\
\bottomrule
\end{tabular}
\end{table}

\begin{figure}[t]
\centering
\includegraphics[width=\linewidth]{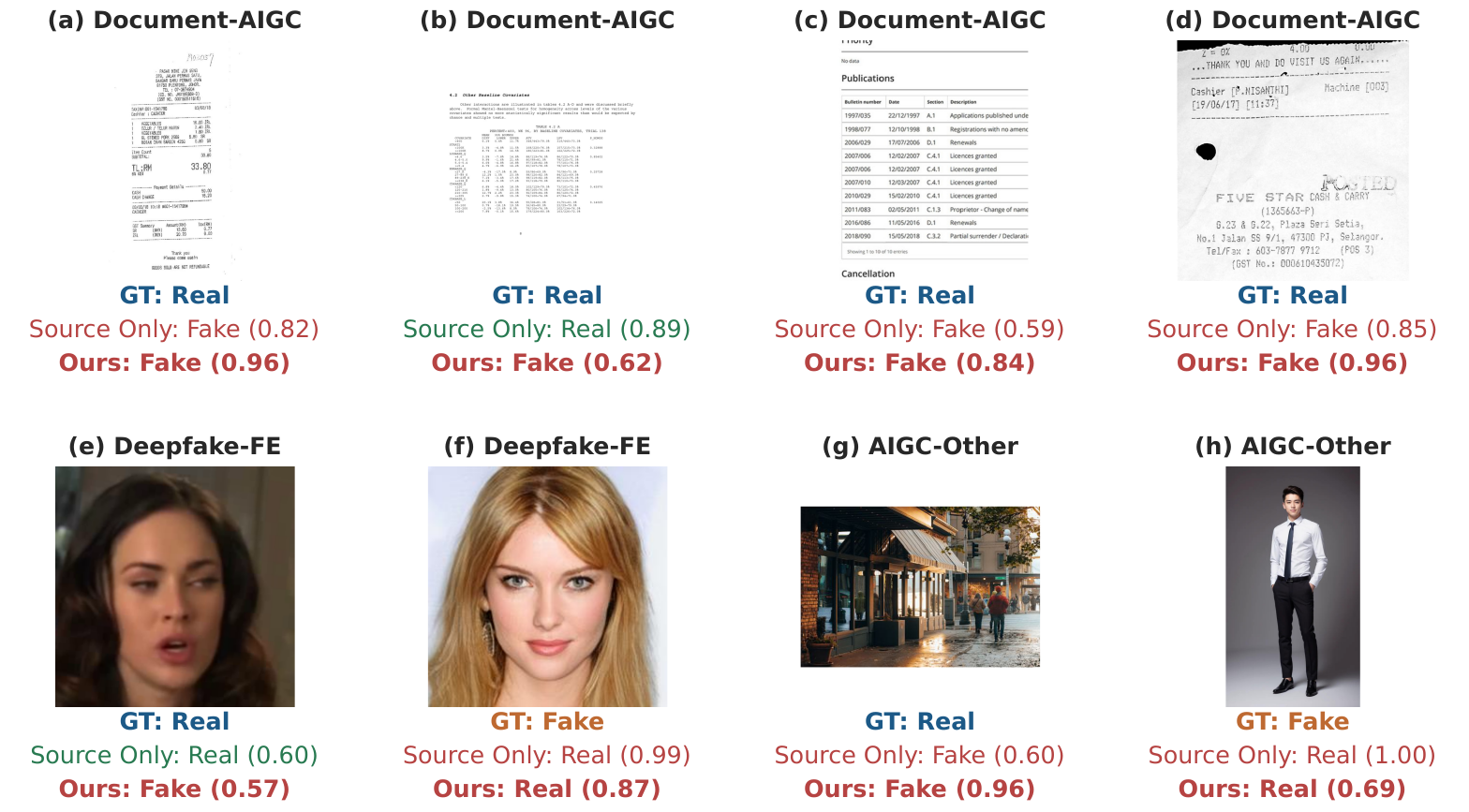}
\caption{\textbf{Qualitative examples of challenging predictions.}
Representative misclassified test images from Document-AIGC and other challenging target domains. Each example reports its ground-truth (GT) label and detector predictions.}
\label{fig:challenging_examples}
\end{figure}

\noindent\textbf{Challenges and Analysis.} 
Table~\ref{tab:doc_aigc_diagnostics} summarizes the diagnostic results for Document-AIGC in the OpenMMSec dataset. The source detector achieves only 54.11\% AUC, with fake-anchor precision of 51.4\%, compared with 78.4\% for real anchors. However, a target-supervised linear probe achieves approximately 87.5\% ACC with the encoder frozen. This indicates that useful discriminative information remains in the target features. GraRA improves AUC to 60.88\%, and graph diffusion increases pseudo-label AUC by 2.33 points. However, the final detection performance remains limited. Unlike fully generated natural images, document forgeries may involve only local edits to text or fields while preserving most of the original layout. These subtle changes may provide weak or sparse forensic cues, making image-level classification more difficult. Other domains show different patterns. AIGC-Other covers various generation methods and gains 20.25 ACC points, but its AUC remains below that of the best competing detector. Deepfake-FE involves edits to facial attributes that may preserve most of the original image. In this domain, AUC increases by 2.60 points, while ACC improves by only 0.27 points. These results show that adaptation gains vary across forgery types and evaluation metrics. Figure~\ref{fig:challenging_examples} further presents representative misclassified examples.

\noindent\textbf{Future Directions.} 
These observations suggest that more reliable semantic guidance is needed under severe domain shifts. Future work could explore adaptive graph propagation when source predictions are unreliable. Region-aware forensic cues could also help detect local edits in documents. In addition, domain-aware calibration may help convert improvements in sample ranking into higher classification accuracy. These extensions could further improve performance on challenging target domains while preserving GraRA's source-free adaptation setting.

\section{Sensitivity and Robustness Analysis}
\label{app:sensitivity_robustness}

\subsection{Sensitivity to Target Class Ratios}
\label{sec:class_ratio_sensitivity}

\begin{figure}[!thp]
    \centering
    \includegraphics[width=\linewidth]{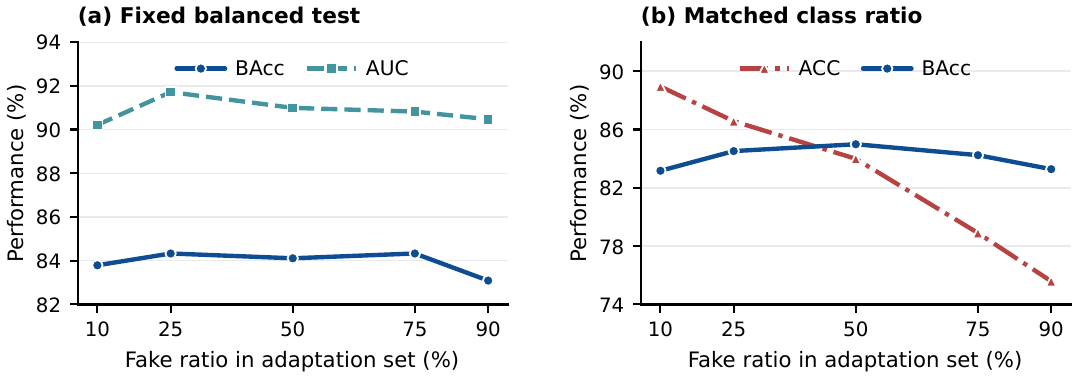}
    \caption{Sensitivity of GraRA to the target class ratio.
    (a) BAcc and AUC on a fixed balanced test set.
    (b) ACC and BAcc under matched adaptation and test ratios.
    Results are averaged across 14 domains and three resamples.}
    \label{fig:class_ratio_sensitivity}
\end{figure}

We vary the fake image fraction in the unlabeled adaptation set from 10\% to 90\% across 14 OpenMMSec target domains, with three resamples per ratio. We evaluate GraRA on a fixed balanced test set and under matched adaptation and test ratios. Target labels are used only for controlled sampling and evaluation. As shown in Fig.~\ref{fig:class_ratio_sensitivity}, GraRA maintains stable balanced accuracy (BAcc) across different ratios, ranging from 83.09\% to 84.33\% under the fixed-test setting and from 83.17\% to 84.99\% under the matched-ratio setting. AUC also remains relatively stable, while ACC varies with the class ratio of the test set. These results show that GraRA maintains similar class-balanced detection performance across different target class ratios.

\begin{figure}[!thp]
\centering
\includegraphics[width=\linewidth]{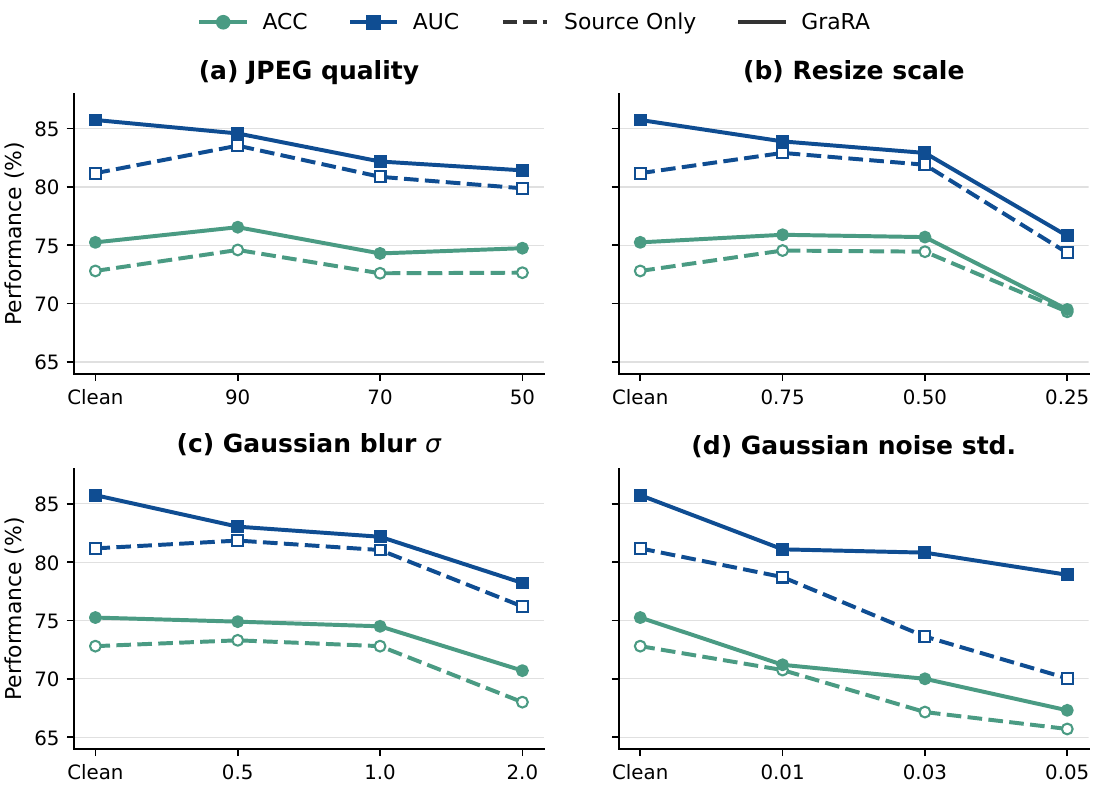}
\caption{Robustness to image perturbations, reported as the unweighted average of dataset-level results on WildDeepfake and WildFake. GraRA is adapted separately on each clean target dataset and kept fixed during perturbed testing. Solid and dashed lines represent GraRA and Source Only, respectively. Circles and squares denote ACC and AUC.}
\label{fig:external_robustness}
\end{figure}

\subsection{Robustness to Image Perturbations}
\label{app:robustness}

We further examine whether the adapted detector remains effective under image degradation. GraRA is adapted once on clean target data and then evaluated under JPEG compression, resizing, Gaussian blur, and noise without further updates. Figure~\ref{fig:external_robustness} shows that GraRA maintains higher AUC than the source detector across all tested perturbations. Its ACC also remains higher in most cases, except under the strongest resizing, where both methods fall below 70\%. Performance generally declines for both methods as blur and noise increase. These results suggest that GraRA improves robustness to common image perturbations, although severe degradation still reduces detection performance.

\begin{table}[!thp]
\centering
\caption{Target-time efficiency comparison. Params: optimized scalar parameters. Time: seconds. Mem.: peak allocated GPU memory (GB). Lat.: milliseconds per image. LH: full linear head.}
\label{tab:efficiency}
\setlength{\tabcolsep}{3pt}
\renewcommand{\arraystretch}{1}
\small
\begin{tabular}{lcccc}
\toprule
Method & Params (\#) & Time (s) & Mem. (GB) & Lat. (ms/img) \\
\midrule
TENT~\cite{wang2021tent}  & 102,400 & 104.62 & 7.03 & 18.40 \\
EATA~\cite{niu2022efficient}  & 102,400 & 51.44 & 12.62 & 17.91 \\
LH    & 769 & 51.80 & \textbf{1.41} & 17.53 \\
\rowcolor{cvprblue!10}
\textbf{GraRA} & \textbf{32} & \textbf{45.43} & \textbf{1.41} & \textbf{17.26} \\
\bottomrule
\end{tabular}
\end{table}

\section{Computational Efficiency}
\label{sec:efficiency}

We compare target-time adaptation efficiency on 14 OpenMMSec
domains with a fixed source checkpoint and identical target
partitions. All timing measurements use a single RTX 3090 GPU
in FP32 with batch size 16. This \emph{efficiency benchmark}
uses a different precision and batch-size configuration from
the main feature-extraction implementation.
Adaptation time is measured from the first target forward pass
until the predictor is ready; model and data loading are excluded. Adaptation times are averaged across the 14 target domains. Table~\ref{tab:efficiency} shows that GraRA optimizes only 32 scalars, compared with 769 parameters in the full linear head (LH).
GraRA achieves the shortest adaptation time (45.43\,s), compared
with 51.80\,s for LH, 51.44\,s for EATA, and 104.62\,s for TENT.
It also shares the lowest peak GPU memory usage with LH (1.41\,GB)
and achieves the lowest inference latency (17.26\,ms/image).
The learned residual is merged into the linear head, requiring
no additional inference branch.


\end{document}